\documentclass[10pt,twocolumn,letterpaper]{article}

\usepackage[pagenumbers]{cvpr} 

\usepackage{amsmath,amsfonts,bm}

\def\Figref#1{Figure~\ref{#1}}

\def\eqref#1{equation~\ref{#1}}

\def\1{\bm{1}}

\def\eps{{\epsilon}}

\def\vv{{\bm{v}}}

\def\vz{{\bm{z}}}

\DeclareMathAlphabet{\mathsfit}{\encodingdefault}{\sfdefault}{m}{sl}
\SetMathAlphabet{\mathsfit}{bold}{\encodingdefault}{\sfdefault}{bx}{n}

\def\gL{{\mathcal{L}}}

\def\gP{{\mathcal{P}}}

\def\gV{{\mathcal{V}}}

\newcommand{\gray}[1]{\textcolor{gray}{#1}}

\newcommand{\conditionalcomment}[1]{\if\commenttext1 \else {#1} \fi}
\newcommand{\grayconditionalcomment}[1]{\if\commenttext1 \else \gray{{#1}} \fi}

\renewcommand{\eqref}[1]{Eq.~(\ref{#1})}

\makeatletter
\DeclareRobustCommand\onedot{\futurelet\@let@token\@onedot}
\def\@onedot{\ifx\@let@token.\else.\null\fi\xspace}

\def\eg{\emph{e.g}\onedot} 
\def\ie{\emph{i.e}\onedot}

\def\etal{\emph{et al}\onedot}
\makeatother

\usepackage{url}
\usepackage{xspace}
\usepackage{amssymb}
\usepackage[accsupp]{axessibility}

\usepackage{graphicx}
\usepackage{multirow}
\usepackage{lipsum} 
\usepackage{booktabs}
\usepackage{algorithm}
\usepackage{algpseudocode, algorithmicx}
\usepackage{amssymb}
\usepackage{pifont}
\usepackage{colortbl}
\definecolor{blue(ncs)}{rgb}{0.0, 0.53, 0.74}
\definecolor{green(ncs)}{rgb}{0.0, 0.62, 0.42}
\definecolor{cadmiumorange}{rgb}{0.93, 0.53, 0.18}

\usepackage[dvipsnames]{xcolor}

\definecolor{grey}{rgb}{0.9, 0.9, 0.9}
\newcommand{\ccol}{\cellcolor{grey}}

\usepackage{tabularx}
\usepackage[export]{adjustbox}

\newcommand{\smallbreakparagraph}[1]{\smallbreak \noindent \textbf{#1}}

\makeatletter
\let\OldStatex\Statex
\renewcommand{\Statex}[1][3]{%
  \setlength\@tempdima{\algorithmicindent}%
  \OldStatex\hskip\dimexpr#1\@tempdima\relax}

\definecolor{cvprblue}{rgb}{0.21,0.49,0.74}
\usepackage[pagebackref,breaklinks,colorlinks,citecolor=cvprblue]{hyperref}

\def\paperID{11552}
\def\confName{CVPR}
\def\confYear{2024}

\title{Contrastive Mean-Shift Learning for Generalized Category Discovery}

\author{Sua Choi \quad Dahyun Kang \quad Minsu Cho\vspace{0.15cm}\\
Pohang University of Science and Technology (POSTECH), South Korea\\
{
\small
\small \url{https://cvlab.postech.ac.kr/research/cms}
}
}

\begin{document}
\maketitle

\begin{abstract}
We address the problem of generalized category discovery (GCD) that aims to partition a partially labeled collection of images; only a small part of the collection is labeled and the total number of target classes is unknown.
To address this generalized image clustering problem, 
we revisit the mean-shift algorithm, \ie, a classic, powerful technique for mode seeking, and incorporate it into a contrastive learning framework.
The proposed method, dubbed Contrastive Mean-Shift (CMS) learning, trains an image encoder to produce representations with better clustering properties by an iterative process of mean shift and contrastive update. 
Experiments demonstrate that our method, both in settings with and without the total number of clusters being known, achieves state-of-the-art performance on six public GCD benchmarks without bells and whistles.

\end{abstract}    
\section{Introduction}
Clustering is one of the most fundamental problems in unsupervised learning, which aims to partition instances of a data collection into different groups~\cite{kmeans, kmeanspp, dbscan, slink}.
Unlike the classification problem, it does not assume either predefined target classes or labeled instances in its standard setup. 
However, in a practical scenario, some data instances may be labeled for a subset of target classes so that we can leverage them to cluster all the data instances while also discovering the remaining unknown classes.
The goal of Generalized Category Discovery (GCD)~\citep{vaze2022gcd} is to tackle such a semi-supervised image clustering problem given a small amount of incomplete supervision.

\begin{figure}[t!]
	\centering
	\small
        \vspace{4mm}
    \includegraphics[width=\linewidth]{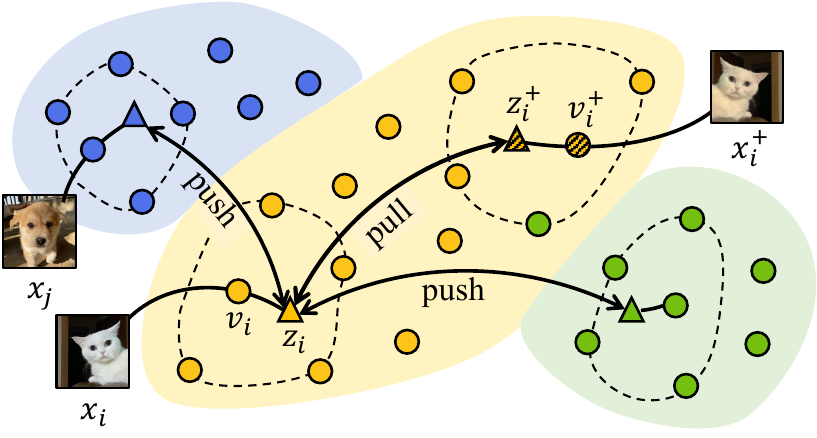}
    \includegraphics[width=1.03\linewidth]{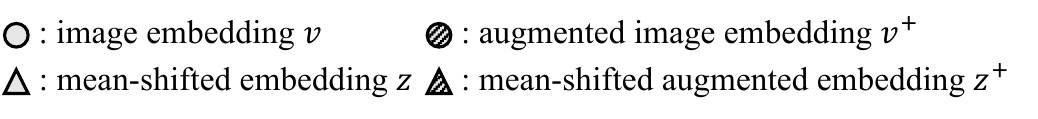}
 \vspace{-7.0mm}
 \caption{\textbf{Contrastive Mean-Shift (CMS) learning.} By integrating mean shift~\cite{fukunaga1975estimation, meanshift} into contrastive learning~\cite{zhuang2019local, chen2020simple}, 
the proposed method learns an embedding space such that the mean-shifted embeddings of identical images $x_{i}$ and $x_{i}^{+}$ draw together and those of distinct images $x_{i}$ and $x_{j}$ push apart from each other. 
}
 
 \vspace{-3mm}
\label{fig:teaser}
\end{figure}



Viewing GCD as a transductive learning problem for semi-supervised clustering, we revisit the mean shift~\cite{fukunaga1975estimation, cheng1995mean, meanshift},
\ie, a classic, powerful technique for mode seeking and clustering analysis.
The mean-shift algorithm assigns each data point a corresponding mode through iterative shifts by kernel-weighted aggregation of neighboring points so as to cluster the data points according to their modes; 
this process is non-parametric and does not require any information about the target clusters, \eg, the number of clusters.

By incorporating the mean shift into a contrastive learning framework~\cite{zhuang2019local, chen2020simple, supervisedcontrastive}, we introduce {\em contrastive mean-shift learning} that induces an embedding space with better clustering properties for GCD.
As illustrated in Fig.~\ref{fig:teaser}, we first perform a single step of mean shift for each image embedding by moving it towards the mean of its neighbors in an embedding space.
We then update the embedding space such that an image $x_{i}$ and its augmented image $x^{+}_{i}$ are drawn closer together while distinct images $x_{i}$ and $x_{j}$ are pushed apart from each other.
These alternative updates with mean-shifted embeddings encourage the image encoder to learn representations with better clustering properties. 
After the encoder is trained, the actual clustering is performed by agglomerative clustering with the learned embeddings.

Prior arts for GCD~\cite{vaze2022gcd, zhang2023promptcal, pu2023dccl, chiaroni2023pim_gcd, wen2023simgcd} typically employ $k$-means clustering~\cite{kmeans, kmeanspp} in validation and testing 
where the ground-truth number of target classes $K$ is often required for stable performance, which is undesirable in practical scenarios. In contrast, our method jointly estimates $K$ during training so that it achieves robust performance without using the ground-truth $K$ in clustering. 

The proposed method is extensively evaluated on the six public GCD datasets~\cite{cifar, geirhos2018imagenet, cub, krause20133scars, maji2013aircraft, tan2019herbarium}, including coarse-grained and fine-grained classification problems, and achieves the state-of-the-art performance on the public GCD benchmarks~\cite{vaze2022gcd, zhang2023promptcal}.
Notably, even when the ground-truth number of target classes $K$ is not used, it shows comparable performance to the state-of-the-art methods that exploit the ground-truth $K$. 

Our contribution can be summarized as follows:
\begin{itemize}
    \item We introduce contrastive mean-shift learning for GCD by incorporating the mean-shift algorithm in a contrastive learning framework.
    \item The proposed method jointly estimates the number of target classes in training and thus achieves robust discovery without using the ground-truth number of target classes.
    \item Extensive experiments and analyses demonstrate that our method outperforms the state-of-the-art methods on several standard GCD benchmarks. 
\end{itemize}

\section{Related work}

\subsection{Generalized Category Discovery (GCD)}
The task of GCD~\cite{vaze2022gcd} 
is generalized from novel category discovery (NCD)~\cite{ncd}; in GCD unlabeled images may come from either known or unknown classes, whereas in NCD all unlabeled images are from unknown classes. 
Existing work for GCD focuses on transferring information from labeled images to unlabeled ones.
A line of work leverages pseudo-labels of the unlabeled images for learning:
DCCL~\citep{pu2023dccl} adopts InfoMap clustering for pseudo-labeling, PromptCAL~\citep{zhang2023promptcal} discovers pseudo-positive samples based on semi-supervised affinity generation, and SimGCD~\citep{wen2023simgcd} adopts a parametric classifier by distilling reliable pseudo-labels.
The other popular approach proposes a semi-supervised learning objective~\cite{cst, zhou2019collaborative, tieredimagenet, s3d}: the seminal work~\cite{vaze2022gcd} employs a supervised contrastive loss with labeled images and a self-supervised contrastive loss, and PIM~\cite{chiaroni2023pim_gcd} adopts bi-level mutual information optimization.

Despite the progress, it is worth noting that most previous work adopts a sequential two-stage framework for estimating the number of classes after training the model. 
Some recent work~\cite{pu2023dccl, zhao2023gmm_gcd} jointly estimates the number of classes during training but requires heavy computation to update pairwise similarities of the entire image collection~\cite{pu2023dccl} or manual hyperparameter tuning for cluster merging and splitting~\cite{zhao2023gmm_gcd}.
Furthermore, most of the work exploits the ground-truth number of classes for model selection~\cite{vaze2022gcd, zhang2023promptcal, pu2023dccl, chiaroni2023pim_gcd} or classifier learning~\cite{wen2023simgcd}.

In contrast, our method shows robust performance even without access to the ground-truth number of classes by efficiently estimating it during training.  
Moreover, our simple learning framework does not involve additional techniques such as the teacher-student framework~\cite{zhang2023promptcal, wen2023simgcd}, pseudo-labeling~\cite{pu2023dccl}, and add-on classifiers~\cite{wen2023simgcd}. 

\subsection{Mean shift}

The mean shift is a non-parametric feature-space analysis technique for locating the maxima, \ie, the modes, of a density function or data distribution~\cite{fukunaga1975estimation,meanshift}. 
It was first introduced by Fukunaga and Hostetler~\cite{fukunaga1975estimation} as a kernel-based approach to estimate the gradient of a density function.
Recursively shifting the data points according to this estimation enables mode-seeking and clustering of data.
Cheng~\cite{cheng1995mean} clarifies the generalized formulation of mean shift across different choices of kernels and algorithms, and demonstrates its convergence over iterations.
A line of work explores diverse variants of the mean-shift algorithm for clustering. For example, Kobayashi and Otsu~\cite{kobayashi2010hypersphere} perform mean-shift clustering on a hypersphere using the von Mises-Fisher distribution~\cite{mardia2000directional}, and Yuan~\etal~\cite{yuan2010agglomerative} suggest an agglomerative mean-shift algorithm for computationally efficient clustering.
The mean shift has been widely studied in computer vision for various applications, \eg, object tracking~\cite{comaniciu2000real, comaniciu2003kernel, kumar2022gridshift, jang2021meanshiftpp}, image segmentation~\cite{kumar2022gridshift, jang2021meanshiftpp, kong2018recurrent}, line segment detection~\cite{bandera2006houghmean}, 
and scene summarization~\cite{cho2012mode}.

\begin{figure*}[t!]
	\centering
	\small
    \includegraphics[width=0.95\linewidth]{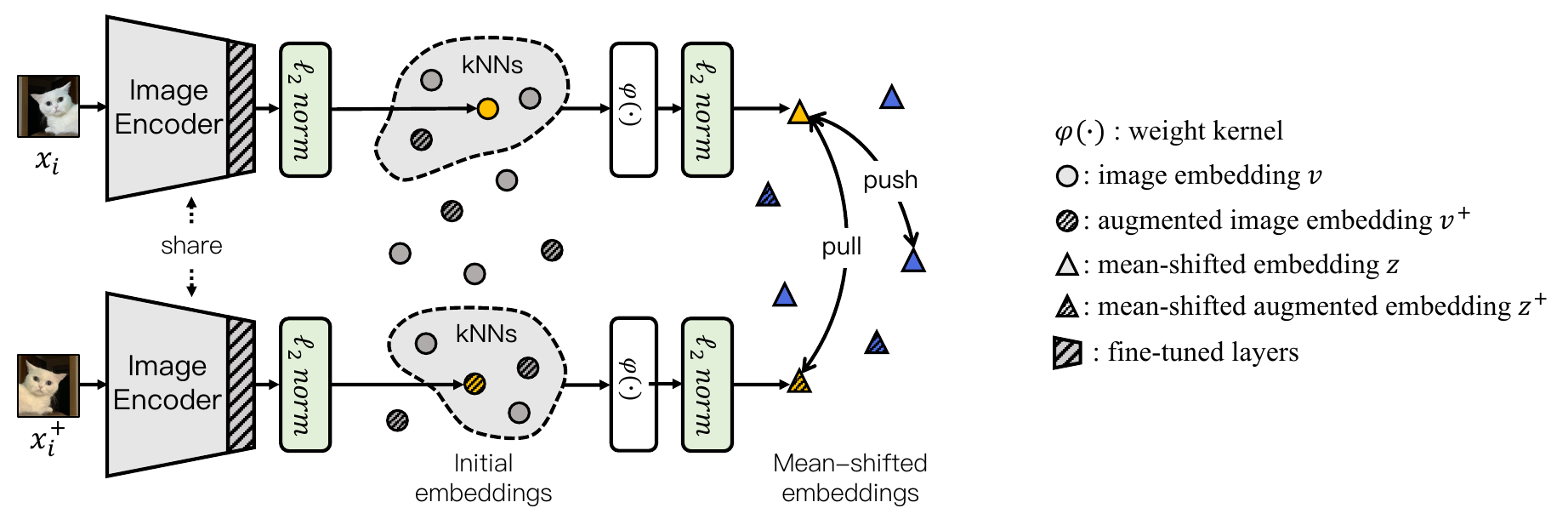}
 \vspace{-1.0mm}
 \caption{
 \textbf{Contrastive Mean-Shift Learning.}
Given a collection of images, each initial image embedding $\vv_i$ from an image encoder takes a single step of mean shift to be $\vz_i$ by aggregating its $k$ nearest neighbors with a weight kernel $\varphi(\cdot)$.
The encoder network is then updated by contrastive learning with the mean-shifted embeddings, which draws a mean-shifted embedding of image $x_{i}$ and that of its augmented image $x_{i}^{+}$ closer and pushes those of distinct images apart from each other. See text for details. 
}

\label{fig:overview}
\end{figure*}

\section{Preliminaries}
\subsection{Problem definition and setup}

Let us denote an image-label pair by $(x, y) \in \mathcal{X} \times \mathcal{Y}$ where $\mathcal{X}$ and $\mathcal{Y}$ represent the set of all possible images and that of all class labels, respectively. We adopt a symbol ${\eps}$ for {\em no label} and represent an {\em unlabeled} image by $(x, \eps)$.

Now consider a partially-labeled image collection $\mathcal{D} = \mathcal{D}_{\text{L}} \cup \mathcal{D}_{\text{UL}}$ that is comprised of a labeled set $\mathcal{D}_{\text{L}} \subset \mathcal{X}\times \mathcal{Y}_\text{KN}$ and an unlabeled set $\mathcal{D}_{\text{UL}} \subset \mathcal{X}\times \{ \eps \}$; $\mathcal{Y}_\text{KN}$ denotes a set of {\em known} classes, $\mathcal{Y}_\text{KN} \subset \mathcal{Y}$.
The task of GCD is then to partition $\mathcal{D}$ into multiple disjoint clusters $\{ \mathcal{C}_c \}_{c=1}^{K}$ such that each cluster $\mathcal{C}_c$ collects all images with the same class from $\mathcal{D}$. 
The total number of clusters, \ie, classes, $K$ is unknown, and a separate validation set  $\mathcal{D}_\text{V}$ is provided~\cite{vaze2022gcd}. 

Note that in the unlabeled set $\mathcal{D}_{\text{UL}}$, some images may not belong to the known classes $\mathcal{Y}_\text{KN}$, \ie, classes in the labeled set.  
We call classes of those images {\em unknown} classes $\mathcal{Y}_\text{UKN}$: $\mathcal{Y}_\text{KN} \cup \mathcal{Y}_\text{UKN} \subset \mathcal{Y}$ and $|\mathcal{Y}_\text{KN} \cup \mathcal{Y}_\text{UKN}| = |\mathcal{Y}_\text{KN}| + |\mathcal{Y}_\text{UKN}|= K$. 
Accordingly, the image collection $\mathcal{D}$ consists of three types of images: known-labeled, known-unlabeled, and unknown-unlabeled. The validation set $\mathcal{D}_\text{V}$ includes a relatively small number of known-labeled and unknown-unlabeled images. For final evaluation, the actual dataset contains ground-truth labels for all images.



\subsection{Mean-shift algorithm}
Given a collection of data points $\mathcal{V}$ in a feature space, the weighted mean $m(\vv_{i})$ of each data point $\vv_{i}$ is calculated over its neighborhood $\mathcal{N}(\vv_{i})\subseteq \mathcal{V}$ as:
\begin{equation}
    m(\vv_{i}) = \frac{\sum_{\vv_j \in \mathcal{N}(\vv_{i})} \varphi(\vv_j - \vv_{i}) \vv_j} {\sum_{\vv_j \in \mathcal{N}(\vv_{i})} \varphi(\vv_j - \vv_{i})}, \label{eq:ms}
\end{equation}
where a kernel function $\varphi(\cdot)$ determines weights for neighbors in estimating the mean. 
The mode of $\vv_{i}$ is sought by iteratively shifting to its weighted mean until convergence. 
The set of data points that converge to the same mode defines the basin of attraction of that mode, and this naturally relates to clustering: the points in the same basin of attraction are associated with the same cluster~\cite{meanshift}.  

The mean shift is characterized by the set of neighbors $\mathcal{N}(\vv_{i})$ and the kernel function $\varphi(\cdot)$. 
In typical setups~\cite{cheng1995mean, meanshift, comaniciu2000real, wu2007mean}, $\mathcal{N}(\vv_{i})$ is defined by a certain radius and $\varphi(\cdot)$ is set to a uniform, Gaussian, or Epanechnikov kernel~\cite{scott2015multivariate}.

\section{Our approach}
We propose \textit{contrastive mean-shift learning} for GCD by integrating the mean shift in the contrastive learning framework.
Figure~\ref{fig:overview} illustrates the overall procedure.  

Given a collection of images with partial labels, we obtain initial image embeddings from a self-supervised encoder network~\cite{dino}, perform a single-step mean shift on each of them using its $k$ nearest neighbors ($k$NNs) (Sec.~\ref{sec:mean_shifted_embedding}), and then update the last layer of the encoder through contrastive learning~\cite{zhuang2019local, chen2020simple, supervisedcontrastive} across the mean-shifted embeddings (Sec.~\ref{sec:learning}). 
After each epoch of training, the number of classes $K$ is estimated by agglomerative clustering~\cite{ward1963hierarchical} and used to measure the clustering accuracy of the snap-shot model on the validation set (Sec.~\ref{sec:clustering}). This update procedure is performed for a sufficient number of epochs, and the best model is selected according to the validation accuracy.  


After training the encoder, we apply multi-step mean shifts on the final embedding space and the final cluster assignment is performed using the number of clusters $K$ estimated in training (Sec.~\ref{sec:iteration}).

\subsection{Mean-shifted embedding}
\label{sec:mean_shifted_embedding}
Given a collection of images $\mathcal{X} = \{x_1, x_2,\cdots, x_N \}$, the images are fed to an image encoder $f$ to generate the corresponding set of $d$-dimensional $l2$-normalized embeddings: 
\begin{equation}
\mathcal{V}= \{\vv_i\}_{i=1}^{N}, \quad \text{where } \;  \vv_i = f(x_{i}). 
\end{equation}
To obtain discriminative initial embeddings without supervision, we use the self-supervised pre-trained encoder, DINO~\cite{dino}; our method is not restricted to a specific encoder.

The \textit{mean-shifted embedding} $\vz_{i}$ is obtained from the initial embedding $\vv_{i}$ using a single-step mean shift similar to Eq.~(\ref{eq:ms}). 
The conventional mean shift typically defines the neighborhood for each data point based on a distance, \ie, radius.
We find that the number of neighbors within a fixed radius varies significantly during the update of the encoder, causing the training phase to be unstable. 
To address the issue, we replace the distance-based NNs with $k$NNs, which greatly improves the stability and is also suitable for parallel computation with GPUs. 
The neighborhood $\mathcal{N}(\vv_{i})$ is thus defined with input $\vv_{i}$ and its $k$NNs: 
\begin{equation}
    \mathcal{N}(\vv_{i}) = \{\vv_{i} \} \cup \mathrm{argmax}^k_{\vv_j \in \gV}  \vv_{i} \cdot \vv_j,
    \label{eq:knn}
\end{equation}
where $\mathrm{argmax}^k_{s \in \mathcal{S} }(\cdot)$ returns a subset of the top-$k$ items that maximizes a target function.

Along with the neighborhood, we design the weight kernel $\varphi(\cdot)$ to put a higher weight on the center, \ie, the query position $\vv_{i}$, compared to its $k$NNs in aggregation:
\begin{align}
    \varphi(\vv) &= 
    \begin{cases}
         1 - \alpha 
                &\text{if } ||\vv|| = 0 \\
          \frac{\alpha}{k} & \text{otherwise},
    \end{cases}
    \label{eq:mean_shifted_embedding_meanvector}
\end{align}
where $\alpha$ denotes a scaling hyperparameter ($\alpha = 0.5$ in our experiment).
This kernel can be interpreted as a simple approximation of a Gaussian kernel with adaptive bandwidth.

The mean-shifted embedding $\vz_{i}$ is obtained by aggregating the neighbor embeddings with the kernel and then $l2$-normalizing it, ensuring that the shifted embedding remains on a unit hypersphere:
\begin{align}
    \vz_{i}
    &= \frac{\sum_{\vv_j \in \mathcal{N}(\vv_{i})} \varphi(\vv_j - \vv_i) \vv_j}{||\sum_{\vv_j \in \mathcal{N}(\vv_{i})} \varphi(\vv_j - \vv_i) \vv_j||}. \quad
    \label{eq:mean_shifted_embedding}
\end{align}

We use these mean-shifted embeddings in contrastive learning to update the encoder, which is described next.



\subsection{Contrastive mean-shift learning}
\label{sec:learning}
The objective of contrastive mean-shift learning is to encourage the model to improve its clustering properties in the mean-shifted embedding space, bringing closer the mean-shifted embeddings of an identical image while pushing apart those of distinct images.
The learning objective consists of two types of terms:
(1) the unsupervised contrastive mean-shift loss $\gL_{\text{CMS}}$ for all images $\mathcal{D}_\text{L} \cup \mathcal{D}_\text{UL}$ and (2) the supervised contrastive loss $\gL_{\text{SC}}$ for the labeled set $\mathcal{D}_\text{L}$.

We apply random image augmentation with cropping, flipping and color jittering~\cite{vaze2022gcd} to all images in a batch and create positive pairs, $x_{i}$ and its augmented version $x^{+}_{i}$ while considering pairs of two distinct images, $x_{i}$ and $x_{j}$, as negative pairs. 
The unsupervised contrastive mean-shift loss is designed to decrease the distance between the mean-shifted embeddings of the positive pair and increase the distance between those of the negative pair. The individual loss term for the mean-shifted embedding $\vz_{i}$ is formulated as: 
\begin{equation}    
    \gL^{(i)}_{\text{CMS}} = -\log\frac{\exp(\vz_{i} \cdot \vz^{+}_{i}/\tau_{u})}{\sum_{j \neq i} \exp(\vz_{i} \cdot \vz_{j}/\tau_{u})}, \label{eqn:loss_unsup}
\end{equation}
where 
$\tau_{u}$ is a hyperparameter for adjusting the temperature.

Similarly, the supervised contrastive learning loss~\cite{supervisedcontrastive, vaze2022gcd} is formed with the labeled images only,
which decreases the distance between the features of the same class and increases the distance between the others according to the given ground-truth class labels:
\begin{equation}    
    \gL^{(i)}_{\text{SC}} = -\frac{1}{|\gP(i)|} \sum_{\substack{p \in \gP(i})} \log\frac{\exp(\vv_{i} \cdot \vv_{p}/\tau_{s})}{\sum_{j \notin \gP(i)}\exp(\vv_{i} \cdot \vv_{j}/\tau_{s})}, \label{eqn:loss_sup}
\end{equation}
where $\gP(i)$ represents the set of image indices for the same class with image $x_i$ in a batch. 

Denoting by $\mathcal{B}$ the set of image indices in a batch and by $\mathcal{B}_{\text{L}}$ its subset for labeled images, 
the overall learning objective combines the two types of individual losses:
\begin{align}
    \gL = \lambda \frac{1}{|\mathcal{B}_{\text{L}}|} \sum_{i \in \mathcal{B}_{\text{L}}} \gL^{(i)}_{\text{SC}} + (1-\lambda) \frac{1}{|\mathcal{B}|} \sum_{i \in \mathcal{B}} \gL^{(i)}_{\text{CMS}},
    \label{eqn:loss}
\end{align}
where $\lambda$ represents the hyperparameter balancing between two types of losses.

We only train the last block of the pre-trained image encoder~\cite{dino} with the projection head, which amounts to 6.3M parameters for learning in our case.

\subsection{Estimating the number of clusters}
\label{sec:clustering}
During training, we also estimate the number of clusters $K$ by measuring the clustering accuracy on the validation set $\mathcal{D}_{\text{V}}$. 
For actual clustering, we use the agglomerative clustering algorithm with the ward linkage criterion~\cite{ward1963hierarchical}, which iteratively merges the closest pair of clusters until it reaches a certain distance threshold or a target number of clusters. At the end of each training epoch, we apply the algorithm to $\mathcal{D}_\text{V}$ and obtain clustering results for different numbers of clusters.
Among them, the highest clustering accuracy, which is measured using the labeled images in $\mathcal{D}_\text{V}$, is recorded as the validation performance at the epoch and the corresponding number of clusters is determined as the estimated number of clusters $K$ at the epoch. 
Once the training process is over, the snapshot model at the epoch with the best validation performance is selected as the final model, and the estimated $K$ at the epoch is determined as the final estimation of the number of clusters. 
This approach allows us to avoid accessing the ground-truth number of classes during both training and validation, in contrast to previous work~\cite{vaze2022gcd, pu2023dccl, zhang2023promptcal, chiaroni2023pim_gcd, wen2023simgcd}.

In the validation setup of GCD, where the number of images is relatively small, \ie, \ $|\mathcal{D}_\text{V}| \ll |\mathcal{D}|$, the proposed method based on agglomerative clustering is significantly more efficient than the commonly used one~\cite{vaze2022gcd} based on $k$-means clustering.
The computational cost of $k$-means clustering increases linearly with the number of data points and the number of target clusters $k$, \ie, $O(k|\mathcal{D}_\text{V}|)$ in our case, while that of agglomerative clustering amounts to $O(|\mathcal{D}_\text{V}|^{2})$. 
However, since the $k$-means clustering requires a restart to produce a different number of clusters, searching for the optimal number of clusters $K$ using the $k$-means requires a large number of trials with increasing $k$, resulting in a prohibitive cost during training in practice. 
In contrast, once a matrix with pairwise linkage distances is computed, agglomerative clustering produces clustering results for all different granularities by sequential merging in a single run, enabling efficient estimation of $K$ during training. 

\floatname{Algorithm}{Training process of \ours}
\renewcommand{\algorithmicrequire}{\textbf{Input:}}
\renewcommand{\algorithmicensure}{\textbf{Output:}}
\begin{algorithm}[t!]
    \caption{Final clustering inference}
    \label{alg:mean-shift-algorithm}
    \begin{algorithmic}[1] 
        \Require set of image embeddings $\mathcal{V}^{(0)} = \mathcal{V}_\text{L} \cup \mathcal{V}_\text{UL}$, 
        \Statex[1] number of nearest neighbors $k$,
        \Statex[1] number of clusters $K$,
        \Statex[1] number of max iterations $T_{\text{max}}$
        \Ensure final clustering assignments $\text{``cluster''}$
        \For{$t \;\; \text{\bf{in}} \;\; \{0, 1, \cdots,  T_{\text{max}} \} $}
            \State $\text{cluster}^{(t)} \gets \text{AgglomerativeClustering}(\mathcal{V}^{(t)}, K)$
            \State $\text{acc}^{(t)}_{\text{L}} \gets         \text{ComputeAccuracy(cluster}^{(t)}_{\text{L}})$
            \\
            \If {$t > 1$}
                \If {$\text{acc}^{(t-2)}_\text{L} \geq \text{max}(\text{acc}^{(t-1)}_\text{L}, \text{acc}^{(t)}_\text{L})$} 
                    \State \Return $\text{cluster}^{(t-2)}$
                \EndIf
            \EndIf  \Comment{Early stopping}
            \\
            \For{$\vv^{(t)}_{i} \text{ \bf{in} } \mathcal{V}^{(t)}$}
                \State $\mathcal{N}(\vv^{(t)}_{i}) \gets \{\vv^{(t)}_{i} \} \cup k\text{NN set of } \vv^{(t)}_{i}$
                \State $\vv^{(t+1)}_{i} \gets$ {\large $   \frac{\sum_{\vv^{(t)}_j \in \mathcal{N}(\vv^{(t)}_{i})} \varphi(\vv^{(t)}_j-\vv^{(t)}_i) \vv^{(t)}_j}{||\sum_{\vv^{(t)}_j \in \mathcal{N}(\vv^{(t)}_{i})} \varphi(\vv^{(t)}_j-\vv^{(t)}_i) \vv^{(t)}_j||} $}
            \EndFor \Comment{Mean-shift (Eqs.~(\ref{eq:knn}-\ref{eq:mean_shifted_embedding}))}            
        
        \EndFor

    \end{algorithmic}
\end{algorithm}


\subsection{Final clustering inference}
\label{sec:iteration}

The encoder learned by contrastive mean-shift learning is used for the final cluster assignment; we extract embeddings of images using the encoder and partition them into $K$ clusters using agglomerative clustering with the estimated $K$.    
To improve the clustering property of the embeddings, we perform multi-step mean shift on the embeddings before agglomerative clustering. 
Algorithm~\ref{alg:mean-shift-algorithm} summarizes the inference process.
Starting from the initial embeddings $\mathcal{V}^{(0)}$ from the learned encoder, we update them to $t$-step mean-shifted embeddings $\mathcal{V}^{(t)}$ until the clustering accuracy on $\mathcal{D}_{\text{L}}$ does not increase for two consecutive iterations.
The final cluster assignment is obtained by performing agglomerative clustering on the multi-step mean-shifted embeddings. 
We observe that the multi-step mean shift brings a substantial performance gain in clustering, as shown in \Figref{fig:meanshift}. 

\section{Experiments}

\begin{table*}[t!]
    \centering
    \tabcolsep=0.15cm
    \resizebox{2.1\columnwidth}{!}{
    \begin{tabular}[b]{p{0.14\textwidth} ccc ccc ccc ccc ccc ccc}
    \toprule
        \multirow{2.5}{*}{Method} &
        \multicolumn{3}{c}{CIFAR100} &
        \multicolumn{3}{c}{ImageNet100} &
        \multicolumn{3}{c}{CUB} &
        \multicolumn{3}{c}{Stanford Cars} &
        \multicolumn{3}{c}{FGVC Aircraft} &
        \multicolumn{3}{c}{Herbarium 19} \\
        \cmidrule(lr){2-4} \cmidrule(lr){5-7} \cmidrule(lr){8-10}
        \cmidrule(lr){11-13} \cmidrule(lr){14-16} \cmidrule(lr){17-19}
        & All & Old & Novel & All & Old & Novel & All & Old & Novel
        & All & Old & Novel & All & Old & Novel & All & Old & Novel\\
    \midrule
    \multicolumn{19}{l}{\textit{\hypertarget{withK}{(a)} Clustering with the ground-truth number of classes $K$ given }} \\
    \midrule
    \small{Agglomerative}~\cite{ward1963hierarchical}$\dagger$ & 56.9 & 56.6 & 57.5 & 73.1 & 77.9 & 70.6 & 37.0 & 36.2 & 37.3 & 12.5 & 14.1 & 11.7 & 15.5 & 12.9 & 16.9 & 14.4 & 14.6 & 14.4 \\
        RankStats+~\cite{Han2020rs+} & 58.2 & 77.6 & 19.3 & 37.1 & 61.6 & 24.8 & 33.3 & 51.6 & 24.2 & 28.3 & 61.8 & 12.1 & 26.9 & 36.4 & 22.2 & 27.9 & 55.8 & 12.8 \\
        UNO+~\cite{fini2021uno} & 69.5 & 80.6 & 47.2 & 70.3 & 95.0 & 57.9 & 35.1 & 49.0 & 28.1 & 35.5 & 70.5 & 18.6 & 40.3 & 56.4 & 32.2 & 28.3 & 53.7 & 14.7 \\
        ORCA~\cite{cao2022orca} & 69.0 & 77.4 & 52.0 & 73.5 & 92.6 & 63.9 & 35.3 & 45.6 & 30.2 & 23.5 & 50.1 & 10.7 & 22.0 & 31.8 & 17.1 & 20.9 & 30.9 & 15.5\\
        GCD~\cite{vaze2022gcd} & 73.0 & 76.2 & 66.5 & 74.1 & 89.8 & 66.3 & 51.3 & 56.6 & 48.7 & 39.0 & 57.6 & 29.9 & 45.0 & 41.1 & 46.9 & 35.4 & 51.0 & 27.0\\
        DCCL~\cite{pu2023dccl} & 75.3 & 76.8 & 70.2 & 80.5 & 90.5 & 76.2 & 63.5 & 60.8 & \textbf{64.9} & 43.1 & 55.7 & 36.2 & - & - & - & - & - & - \\
        PromptCAL~\cite{zhang2023promptcal} & 81.2 & 84.2 & 75.3 & 83.1 & 92.7 & 78.3 & 62.9 & 64.4 & 62.1 & 50.2 & 70.1 & 40.6 & 52.2 & 52.2 & \textbf{52.3} & 37.0 & 52.0 & 28.9\\
        GPC~\cite{zhao2023gmm_gcd} & 77.9 & 85.0 & 63.0 & 76.9 & 94.3 & 71.0 & 55.4 & 58.2 & 53.1 & 42.8 & 59.2 & 32.8 & 46.3 & 42.5 & 47.9 & - & - & - \\
        SimGCD~\cite{wen2023simgcd} & 80.1 & 81.2 & \textbf{77.8} & 83.0 & 93.1 & 77.9 & 60.3 & 65.6 & 57.7 & 53.8 & 71.9 & 45.0 & 54.2 & 59.1 & 51.8 & \textbf{44.0} & \textbf{58.0} & \textbf{36.4} \\
        PIM~\cite{chiaroni2023pim_gcd} & 78.3 & 84.2 & 66.5 & 83.1 & 95.3 & 77.0 & 62.7 & 75.7 & 56.2 & 43.1 & 66.9 & 31.6 & - & - & - & 42.3 & 56.1 & 34.8\\
        \ccol Ours & \ccol \textbf{82.3} & \ccol \textbf{85.7} & \ccol 75.5 & \ccol \textbf{84.7} & \ccol \textbf{95.6} & \ccol \textbf{79.2} & \ccol \textbf{68.2} & \ccol \textbf{76.5} & \ccol 64.0 & \ccol \textbf{56.9} & \ccol \textbf{76.1} & \ccol \textbf{47.6} & \ccol \textbf{56.0} & \ccol \textbf{63.4} & \ccol \textbf{52.3} & \ccol 36.4 & \ccol 54.9 & \ccol 26.4 \\
    \midrule  
    \multicolumn{19}{l}{\textit{ \hypertarget{withoutK}{(b)} Clustering without the ground-truth number of classes $K$ given }} \\
    \midrule
     \small{Agglomerative}~\cite{ward1963hierarchical}$\dagger$ & 56.9 & 56.6 & 57.5 & 72.2 & 77.8 & 69.4 & 35.7 & 33.3 & 36.9 & 10.8 & 10.6 & 10.9 & 14.1 & 10.3 & 16.0 & 13.9 & 13.6 & 14.1 \\
        GCD~\cite{vaze2022gcd} & 70.8 & 77.6 & 57.0 & 77.9 & 91.1 & 71.3 & 51.1 & 56.4 & 48.4 
                                & 39.1 & 58.6 & 29.7 & - & - & - & 37.2 & 51.7 & 29.4 \\
        GPC~\cite{zhao2023gmm_gcd} & 75.4 & \textbf{84.6} & 60.1 & 75.3 & 93.4 & 66.7 & 52.0 & 55.5 & 47.5 &
                                    38.2 & 58.9 & 27.4 & 43.3 & 40.7 & 44.8 & 36.5 & 51.7 & 27.9 \\
        PIM~\cite{chiaroni2023pim_gcd} & 75.6 & 81.6 & 63.6 & \textbf{83.0} & 95.3 & \textbf{76.9} & 62.0 &                                   \textbf{75.7} & 55.1 & 42.4 & 65.3 & 31.3 & - & - & - & \textbf{42.0} & 55.5 & \textbf{34.7} \\
        \ccol Ours & \ccol \textbf{79.6} & \ccol 83.2 & \ccol \textbf{72.3} & \ccol 81.3 & \ccol \textbf{95.6} & \ccol 74.2 & \ccol \textbf{64.4} & \ccol 68.2 & \ccol \textbf{62.4} & \ccol \textbf{51.7} & \ccol \textbf{68.9} & \ccol \textbf{43.4} & \ccol \textbf{55.2} & \ccol \textbf{60.6} & \ccol \textbf{52.4} & \ccol 37.4 & \ccol \textbf{56.5} & \ccol 27.1 \\
    \bottomrule
    \end{tabular}%
    }
    \vspace{-5mm}
    \caption{Comparison with the state of the arts on GCD, 
    evaluated \textit{with} or \textit{without} the GT $K$ for clustering. $\dagger$ denotes reproduced results. } \vspace{-1mm}
    \label{table:gcd}
\end{table*}%

\begin{table*}[t!]
    \centering
    \tabcolsep=0.12cm
    \resizebox{2.1\columnwidth}{!}{
    \begin{tabular}[b]{p{0.13\textwidth} c ccc ccc ccc ccc ccc ccc}
    \toprule
        \multirow{2.5}{*}{Method} &
        \multirow{2.5}{*}{Known $K$} &
        \multicolumn{3}{c}{CIFAR100} &
        \multicolumn{3}{c}{ImageNet100} &
        \multicolumn{3}{c}{CUB} &
        \multicolumn{3}{c}{Stanford Cars} &
        \multicolumn{3}{c}{FGVC Aircraft} &
        \multicolumn{3}{c}{Herbarium 19} \\
        \cmidrule(lr){3-5} \cmidrule(lr){6-8} \cmidrule(lr){9-11}
        \cmidrule(lr){12-14} \cmidrule(lr){15-17} \cmidrule(lr){18-20}
        & & All & Old & Novel & All & Old & Novel & All & Old & Novel
        & All & Old & Novel & All & Old & Novel & All & Old & Novel\\
    \midrule
        GCD~\cite{vaze2022gcd} & \ding{51} & 70.1 & 76.8 & 43.5 & 79.7 & 92.7 & 66.7 & 57.5 & 64.5 & 50.6 & -& -& -& -& -&-&-&-&-\\
        ORCA~\cite{cao2022orca} & \ding{51} & 77.7 & 83.6 & 53.9 & 81.3 & 94.5 & 68.0 & 40.7 & 61.2 & 20.2 \\
        PromptCAL~\cite{zhang2023promptcal}$\dagger$ & \ding{51} & \textbf{81.6} & \textbf{85.3} & \textbf{66.9} & 84.8 & 94.4 & 75.2 & 62.4 & 68.1 & 56.8 & 49.1 & 63.1 & 35.7 & 50.1 & 56.7 & 43.4 & 40.8 & 44.7 & 36.7\\
        \ccol Ours & \ccol \ding{51} & \ccol 80.7 & \ccol 84.4 & \ccol 65.9 & \ccol \textbf{85.7} & \ccol \textbf{95.7} & \ccol \textbf{75.8} & \ccol \textbf{69.7} & \ccol \textbf{76.5} & \ccol \textbf{63.0} & \ccol 57.8 & \ccol 75.2 & \ccol \textbf{41.0} & \ccol 53.3& \ccol \textbf{62.7} & \ccol 43.8 & \ccol \textbf{46.2} & \ccol \textbf{53.0} & \ccol \textbf{38.9} \\
        \ccol Ours & \ccol & \ccol 80.5 & \ccol 84.5 & \ccol 64.4 & \ccol 84.2 & \ccol 95.6 & \ccol 72.9 & \ccol 69.0 & \ccol 76.4 & \ccol 61.7 & \ccol \textbf{57.9} & \ccol \textbf{75.6} & \ccol 40.8 & \ccol \textbf{53.8} & \ccol 62.6 & \ccol \textbf{44.9} & \ccol 42.4 & \ccol 53.5 & \ccol 30.7 \\
    \bottomrule
    \end{tabular}%
    }    \vspace{-5mm}
    \caption{Comparison of our model and the state-of-the-art models on the inductive GCD setup on six datasets. The performance of \cite{vaze2022gcd, cao2022orca} is taken from PromptCAL.
    We reproduced
    PromptCAL on Stanford Cars, FGVC Aircraft and Herbarium19 with its official implementation.
} \vspace{-3mm}
    \label{table:inductivegcd}
\end{table*}%

\subsection{Experimental setup}
\smallbreakparagraph{Datasets.}
We evaluate our method on six image classification benchmarks:
two standard datasets,  CIFAR100~\citep{cifar} and ImageNet100~\citep{russakovsky2015imagenet, geirhos2018imagenet}, and four fine-grained datasets, CUB-200-2011~\citep{cub}, Stanford Cars~\citep{krause20133scars}, FGVC Aircraft~\citep{maji2013aircraft}, and Herbarium19~\citep{tan2019herbarium}.
To divide target classes into known and unknown class sets, we follow the splits of the Semantic Shift Benchmark (SSB)~\citep{vaze2021ssb} for CUB, Stanford Cars, and FGVC Aircraft while using the splits from the work of~\citep{vaze2022gcd} for the others.
In the CIFAR100 benchmark, 80\% of the classes are set to known classes, and in the other benchmarks, 50\% of the classes are set to known classes.
The labeled set $\mathcal{D}_{\text{L}}$ contains 50\% of known-class images for all benchmarks.
For details of data splits, see Table~\ref{table:split} in Appendix Sec.~\ref{sec:detail_dataset}

\smallbreakparagraph{Training details.}
We use the pre-trained DINO ViT-B/16~\cite{dino, vit} with a projection head as our image encoder, following the existing methods~\cite{vaze2022gcd, zhang2023promptcal, pu2023dccl}, for fair comparison; the projection head consists of three consecutive pairs of a 2,048-dimensional linear layer followed by GeLU activation~\cite{gelu}.
Note that the last layer of DINO and the projection head are trained in learning while all the other parts are frozen. 
We used a single RTX-3090 for all experiments. To reduce the computational cost of $k$NN retrieval, we set the output dimension of the projection head to $d$ to 768, unlike other methods typically setting it to 65,536. 
We set $k=8$ for the $k$NN retrieval and the $k$NN embeddings are detached so that gradients flow only through the query embedding in training.  
The image embeddings are all renewed at the beginning of each training epoch with the updated image encoder.
The kernel scaling hyperparameter $\alpha$ is set to be 0.5.
The temperature $\tau_{u}$ and learning rate are set to be 0.3 and 0.01 for coarse-grained benchmarks and 0.25 and 0.05 for fine-grained benchmarks.
The hyperparameters for the learning objectives are set following \cite{vaze2022gcd}.
For example, $\tau_{s}$ and $\lambda$ is set to 0.07 and 0.35, respectively.
The SGD optimizer~\cite{ruder2016sgd} is used with a batch size of 128 and a weight decay of 0.00005.



\smallbreakparagraph{Evaluation.}
The performance is evaluated by clustering the entire image collection $\mathcal{D}$ and measuring the accuracy on $\mathcal{D}_{\text{UL}}$.
Following \citep{vaze2022gcd}, the accuracy is measured by matching the assignments with ground-truth labels by the Hungarian optimal matching~\cite{kuhn1955hungarian}, based on the number of intersected instances between each pair of classes.
The unpaired classes are considered incorrect predictions, while the instances of the most dominant class within each ground-truth cluster are considered correct when calculating the accuracy.

The accuracy is reported on ``All'' unlabeled data as well as the accuracy on those of known and unknown classes, denoted by ``Old'' and ``Novel'' in tables, respectively.
We evaluate the performance using the output of the iterative inference process. 
We report the accuracy with the estimated $K$ as well as the ground-truth $K$ to compare the previous work that assumes the ground-truth number of target classes is available during the evaluation.

\subsection{Main results}
\smallbreakparagraph{Evaluation on GCD.}
Table~\ref{table:gcd} presents a comparison on the GCD setup in both coarse-grained and fine-grained benchmarks with or without the ground-truth (GT) number of classes $K$.
In Table~\ref{table:gcd}~\hyperlink{withK}{(a)}, we compare our method with the state-of-the-art methods, all evaluated with GT $K$.
Note that the GT $K$ is only used for evaluation purpose in our case, and not for model selection during training.
We present an agglomerative clustering~\cite{ward1963hierarchical} baseline with the pre-trained DINO not trained further.
The other state-of-the-art methods adopt semi-supervised $k$-means clustering, where the $K$ centroids are initialized by the labeled data with the GT $K$.
Our method outperforms existing methods and achieves state-of-the-art performance on five out of six datasets.
The performance gain is particularly significant on CUB and Standard Cars benchmarks, with 4.7\%p and 3.1\%p higher accuracy, respectively.
Compared to the training-free agglomerative clustering baseline, the performance gain signifies the efficacy of the contrastive mean-shift learning.
Overall, our method gains notable performance on both Old and Novel classes, which implies that the knowledge acquired from known classes is successfully transferred to unknown classes through the use of nearest-neighbor embeddings according to the input query.
\label{sec:experiments_main_gcd}

In Table~\ref{table:gcd}~\hyperlink{withoutK}{(b)}, we present the comparison of ours and the state of the arts on the same setup with Table~\ref{table:gcd}~\hyperlink{withK}{(a)} but without having the GT number of classes $K$ known for clustering.
For Vaze~\etal~\cite{vaze2022gcd}, we borrow the results from the work of PIM~\cite{chiaroni2023pim_gcd}.
Our method shows outstanding performance in most scenarios even though it does not access to the GT $K$ in both training and testing.
Our method is even superior to the state-of-the-art methods measured with the known value of $K$ on CUB and FGVC Aircraft.
The results show that our $K$-estimation process incorporated in the training phase performs effectively with no significant performance drop compared to the known-$K$ counterparts.

\begin{table*}[h!]
\begin{minipage}{0.67\linewidth}
\centering
    \scalebox{0.77}{ 
    \tabcolsep=0.12cm
    \begin{tabular}[b]{l cc cc cc cc cc cc}
    \toprule
        \multirow{2.5}{*}{Method} &
        \multicolumn{2}{c}{CIFAR100} &
        \multicolumn{2}{c}{ImageNet100} &
        \multicolumn{2}{c}{CUB} &
        \multicolumn{2}{c}{Stanford Cars} &
        \multicolumn{2}{c}{FGVC Aircraft} &
        \multicolumn{2}{c}{Herbarium 19} \\
        \cmidrule(lr){2-3} \cmidrule(lr){4-5} \cmidrule(lr){6-7} \cmidrule(lr){8-9} \cmidrule(lr){10-11} \cmidrule(lr){12-13} 
         & K & Err(\%) &  K & Err(\%) & K & Err(\%)  & K & Err(\%) &  K & Err(\%) & K & Err(\%) \\
    \midrule
        Ground truth & 100 & - & 100 & - & 200 & - & 196 & - & 100 & - & 683 & - \\
        GCD~\cite{vaze2022gcd} & 100 & 0 & 109 & 9  & 231 & 15.5 & 230 & 17.3 & - & -  & 520 & 23.8 \\
        DCCL~\cite{pu2023dccl} & 146 & 46 & 129 & 29& 172 & 9 & 192 & 0.02 & - & - & - & - \\
        PIM~\cite{chiaroni2023pim_gcd} & 95 & 5 & 102 & 2 & 227 & 13.5 & 169 & 13.8 & - & - & 563 & 17.6 \\
        GPC~\cite{zhao2023gmm_gcd} & 100 & 0 & 103 & 3 & 212 & 6 & 201 & 0.03 & - & - & - & - \\
    \midrule
        Ours & 95 & 5 & 116 & 16 & 168 & 16 & 156 & 20.4 & 90 & 10 & 622 & 8.9 \\
        Ours* & 97 & 3 & 116 & 16 & 170 & 15 & 156 & 20.4 & 98 & 2 & 666 & 2.5 \\
    \bottomrule
    \end{tabular}%
}
\vspace{-2mm}
\caption{
    Estimated number and an error rate of a class number $K$
}
\label{table:ablation_kestimation}
\end{minipage}
\hspace{1mm}
\begin{minipage}{0.32\linewidth}
    \centering
    \vspace{2mm}
    \includegraphics[width=0.93\linewidth]{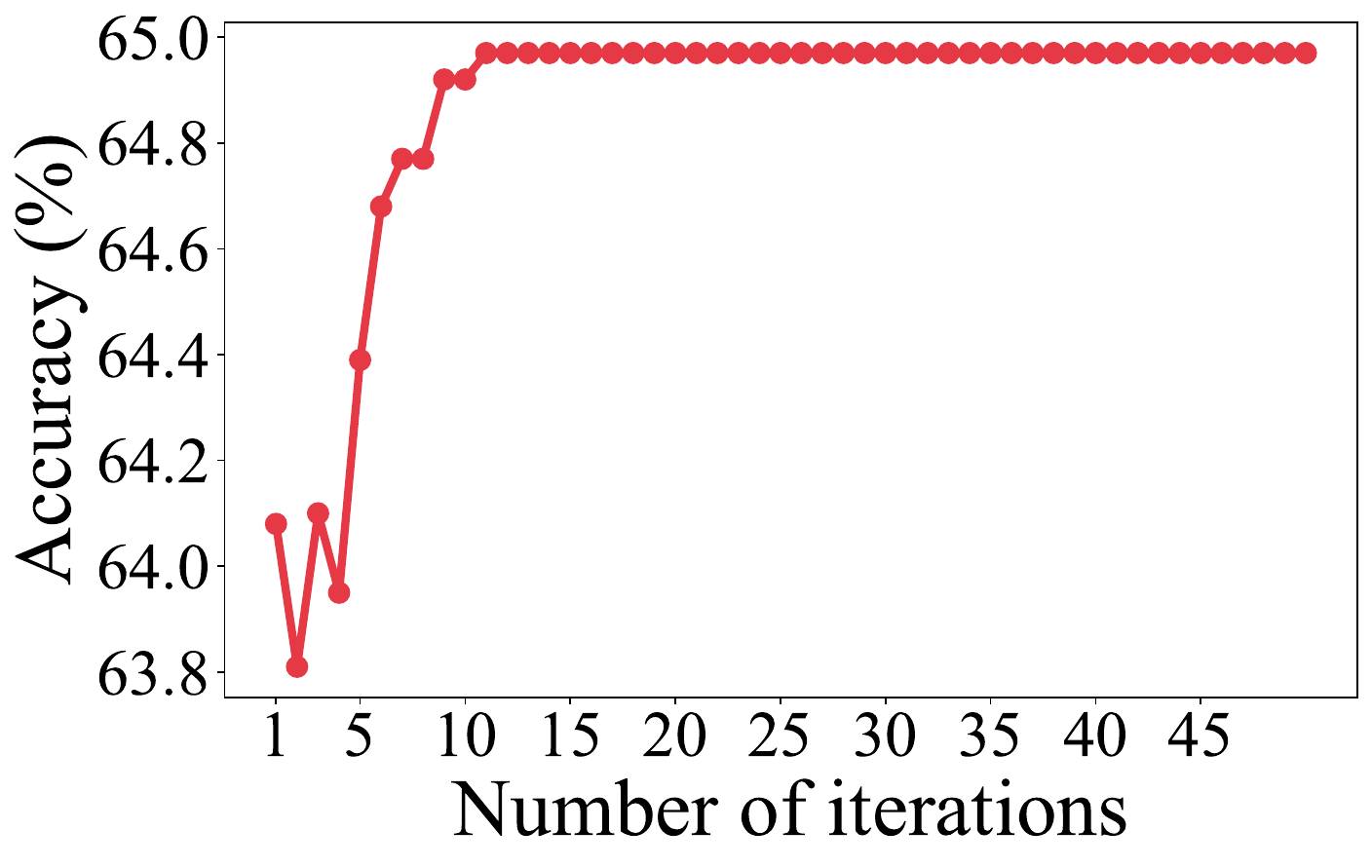}
    \vspace{-3mm}
    \captionof{figure}{Clustering accuracy over mean-shift iterations on CUB.} 
    \label{fig:meanshift}
\end{minipage}
\end{table*}%

\begin{table*}[t!]
        \vspace{-3mm}
        \centering
        \tabcolsep=0.22cm
        \scalebox{0.76}{%
        \begin{tabular}[t]{l  c c c | ccc ccc ccc ccc}
        \toprule
             &
            training &
            \multicolumn{2}{c|}{inference} &
            \multicolumn{3}{c}{CIFAR100}&
            \multicolumn{3}{c}{ImageNet100} &
            \multicolumn{3}{c}{CUB}&
            \multicolumn{3}{c}{Stanford Cars} \\
            \cmidrule(lr){2-2} \cmidrule(lr){3-4} \cmidrule(lr){5-7} \cmidrule(lr){8-10} \cmidrule(lr){11-13} \cmidrule(lr){14-16} 
             & CMS & SSK & IMS & All & Old & Novel & All & Old & Novel & All & Old & Novel & All & Old & Novel\\
        \midrule
            (1) & & \checkmark &  & 71.5 & 77.3 & 60.1 & 74.1 & 89.8 & 66.3 & 51.2 & 49.2 & 52.2 & 37.9 & 57.8 & 28.3 \\
            (2) &  & & \checkmark & 71.6 & 77.3 & 60.0 & 80.3 & 91.7 & 74.6 & 58.7 & 62.0 & 57.1 & 40.8 & 54.5 & 34.2 \\
            (3) & \textbf{\checkmark} & \checkmark & & 81.1 & 85.6 & 72.1 & 83.4 & \textbf{95.8} &77.2 & 66.7 & 75.3 & 62.5 & 54.5 & 76.4 & 43.9 \\
            (4) & \textbf{\checkmark} & & $\diamond$ (1-step) & 80.1 & \textbf{86.0} & 68.4 & 84.1 & 95.6 & 78.3 & \textbf{68.2} & 76.4 & \textbf{64.1} & 56.1 & 74.6 & 47.1 \\
            \ccol (5) & \ccol \textbf{\checkmark} & \ccol & \ccol \textbf{\checkmark} & \ccol \textbf{82.3} & \ccol 85.7 & \ccol \textbf{75.5} & \ccol \textbf{84.7} & \ccol 95.6 & \ccol \textbf{79.2} & \ccol \textbf{68.2} & \ccol \textbf{76.5} & \ccol 64.0 & \ccol \textbf{56.9} & \ccol \textbf{76.1} & \ccol \textbf{47.6} \\
        \bottomrule
        \end{tabular}%
        }
        \vspace{-2mm}
        \caption{Effectiveness of each component of our method. SSK denotes semi-supervised $k$-means clustering and IMS iterative mean-shift.}
        \label{tab:ablation_method}
\end{table*}

\begin{table*}[t!]
    \centering
    \tabcolsep=0.16cm
    \resizebox{2.1\columnwidth}{!}{ 
    \begin{tabular}[b]{l ccc ccc ccc ccc ccc ccc}
    \toprule
        \multirow{2.5}{*}{Kernel} &
        \multicolumn{3}{c}{CIFAR100} &
        \multicolumn{3}{c}{ImageNet100} &
        \multicolumn{3}{c}{CUB} &
        \multicolumn{3}{c}{Stanford Cars} &
        \multicolumn{3}{c}{FGVC Aircraft} &
        \multicolumn{3}{c}{Herbarium 19} \\
        \cmidrule(lr){2-4} \cmidrule(lr){5-7} \cmidrule(lr){8-10}
        \cmidrule(lr){11-13} \cmidrule(lr){14-16} \cmidrule(lr){17-19}
        & All & Old & Novel & All & Old & Novel & All & Old & Novel
        & All & Old & Novel & All & Old & Novel & All & Old & Novel\\
    \midrule
        Uniform & 75.2 & 77.6 & 70.5 & 76.9 & 92.0 & 69.3 & 53.5 & 59.4 & 50.6 & 34.3 & 54.7 & 24.5 & 39.9 & 43.6 & 38.1 & 36.0 & \textbf{55.8} & 25.4 \\
        Gaussian & 72.2 & 80.6 & 55.4 & 67.2 & 94.9 & 53.2 & 45.5 & 43.2 & 46.6 & 34.9 & 56.2 & 24.5 & 36.6 & 36.7 & 36.6 & 25.6 & 38.4 & 18.8 \\
        \ccol Ours & \ccol \textbf{82.3} & \ccol \textbf{85.7} & \ccol \textbf{75.5} & \ccol  \textbf{84.7} & \ccol \textbf{95.6} & \ccol \textbf{79.2} & \ccol \textbf{68.2} & \ccol \textbf{76.5} & \ccol \textbf{64.0} & \ccol \textbf{56.9} & \ccol \textbf{76.1} & \ccol \textbf{47.6} & \ccol \textbf{56.0} & \ccol \textbf{63.4} & \ccol \textbf{52.3} & \ccol \textbf{36.4} & \ccol 54.9 & \ccol \textbf{26.4} \\
    \bottomrule
    \end{tabular}%
    }
    \vspace{-6.5mm}
    \caption{Comparison of ours and different mean-shift kernels on GCD.}
    \label{table:gcd_uniform_gaussian_kernel}
    \vspace{-4mm}
\end{table*}%

\smallbreakparagraph{Evaluation on inductive GCD.}
We also compare the clustering results on the inductive GCD setup presented in PromptCAL~\cite{zhang2023promptcal}.
Contrary to the \textit{transductive} GCD setup in Table~\ref{table:gcd}, the inductive GCD setup evaluates the performance on the \textit{unseen test set}.
In this setup, a subset of the training set is utilized for validation, and the labeled images from the validation set are used to verify the termination condition during the iterative inference process.
The comparison is presented in Table~\ref{table:inductivegcd}, where our method exhibits superior performance in the inductive category discovery scenario as well.
We also report the performance measured with the estimated number of classes $K$, which is more practical as it assumes both unseen data and unknown classes in a test set.
Our method shows comparable performance without the ground-truth $K$ than the other models.
The result demonstrates that incorporating neighborhood embeddings enhances image clustering by ensuring consistency among relevant images, thus enabling generalization to discover novel classes.

\smallbreakparagraph{Estimated number of clusters.}
Table~\ref{table:ablation_kestimation} shows the comparison of the ground-truth $K$, ours, and others reported by Vaze~\etal, DCCL, PIM, and GPC.
Among these baselines, DCCL and GPC jointly estimate the number of classes during training, 
while the others post-estimate $K$ after training as done in \cite{vaze2022gcd}.
Our method estimates class number on par with others \textit{without exploiting any dataset-specific hyperparameters}.
When utilizing the ground-truth $K$ during validation as with other baselines (Ours*), the estimates become more accurate. 
Note that the value is estimated on the validation set, which is relatively small and hence computationally efficient for clustering compared to the previous work~\cite{vaze2022gcd, pu2023dccl, chiaroni2023pim_gcd, zhao2023gmm_gcd} which uses the entire image collection.



\subsection{Ablation study}
\smallbreakparagraph{Performance over mean-shift iterations.}
In Figure~\ref{fig:meanshift}, we examine the clustering accuracy of our method over iterations during the inference phase on CUB benchmark.
The iteration 1 indicates the clustering accuracy using the features extracted from the trained encoder, which is learned to transform a given image to a probable shifted-feature space.
The result demonstrates that iterating the mean shift during inference leads to an additional performance gain without further training.
We empirically observe that the accuracy saturates beyond a certain optimal number of iterations, akin to the behavior of the conventional mean-shift algorithm.
This result shows that iterative mean shift based on $k$NNs roughly approximates the mean-shift kernels that guarantee convergence.

\smallbreakparagraph{Effect of each proposed component.}
Table~\ref{tab:ablation_method} shows the ablation of CMS learning (Sec.~\ref{sec:learning}) and Iterative Mean Shift (IMS, Sec.~\ref{sec:iteration}) for final clustering inteference.
For training, we examine the effect of the embedding without mean shift, \ie, equivalent to the embedding in Vaze~\etal~\cite{vaze2022gcd}.
At inference, semi-supervised $k$-means clustering (SSK)~\cite{vaze2022gcd}, single-step mean shift, and IMS are compared.
Comparing (1) \textit{vs} (3) and (2) \textit{vs} (5), we observe that CMS learning boosts performance significantly.
After training, IMS brings additional gains at inference when comparing (1) \textit{vs} (2) and (3) \textit{vs} (5), plus recursive iterations: (4) \textit{vs} (5).
The final model (5) outperforms others with the combined gain of each proposed component.

\smallbreakparagraph{Comparison with different mean-shift kernels.}
We validate our $k$NN-based mean-shift learning framework by replacing the neighborhood criterion with uniform and Gaussian kernels, which are commonly used for mean shift.
For implementation details, please refer to Appendix~\ref{sec:uniform_gaussian}.
As shown in Table~\ref{table:gcd_uniform_gaussian_kernel}, the performance significantly deteriorates with both kernels.
For constantly updating embedding spaces, it is tricky to set a fixed distance radius for mean-shift kernels.
Noticeably, we observe that the fixed-radius kernels tend to blur the embedding space by incorporating more neighbors within its radius as training progresses. 
In other words, this make irrelevant embeddings to be involved over training, eventually leading the model to produce indistinguishable image embeddings to each other.
Notice that the performance gap with ours (with $k$NN) is larger on the small-scale benchmarks: CUB, Stanford Cars, and FGVC Aircraft, 
which might be more sensitively influenced by the kernel parameters on less data.
Through this, we validate that $k$NN retrieval performs as a less brittle and more GPU-friendly approximation of a Gaussian kernel for feature learning, which adjusts the radius of the mean-shift kernel as the embedding space is updated.


\begin{table}[t]
    \centering
    \tabcolsep=0.148cm
    \scalebox{0.87}{%
    \begin{tabular}[t]{l c c ccc ccc}
    \toprule
        & 
        &
        &
        \multicolumn{3}{c}{ImageNet100} &
        \multicolumn{3}{c}{CUB} \\
        \cmidrule(lr){4-6} \cmidrule(lr){7-9} 
       & $\mathcal{N}(\vv_i)$ & $\varphi(\cdot)$ & All & Old & Novel & All & Old & Novel \\
    \midrule
        (1) & $k$NN & attention & 82.2 & 95.2 & 75.7 & 63.6 & 70.1 & 60.3 \\
        (2) & random & mean & 82.6 & 75.0 & 76.3 & 58.1 & 65.4 & 54.5 \\
        \ccol (3) & \ccol $k$NN & \ccol mean & \ccol \textbf{84.7} & \ccol \textbf{95.6} & \ccol \textbf{79.2} & \ccol \textbf{68.2} & \ccol \textbf{76.5} & \ccol \textbf{64.0} \\
    \bottomrule
    \end{tabular}%
}
\vspace{-2.5mm}
\caption{
    Comparison with different feature aggregation methods}
\vspace{-3mm}
\label{table:ablation_attn}
\end{table}%

\begin{table}[t]
    \centering
    \tabcolsep=0.16cm
    \scalebox{0.87}{%
    \begin{tabular}[t]{l l l l}
    \toprule
        method &
        CIFAR100 &
        ImageNet100 &
        CUB \\
    \midrule
       Vaze~\etal~\cite{vaze2022gcd} & 54.8 & 74.1 & 27.9 \\
       \ccol Ours & \ccol \textbf{60.5} (+5.7\%p) & \ccol \textbf{77.9} (+3.8\%p) & \ccol \textbf{32.4} (+4.5\%p)\\
    \bottomrule
    \end{tabular}%
}
\vspace{-2.5mm}
\caption{
   Comparison of two methods on the unsupervised setup
}
\vspace{-3mm}
\label{table:ablation_unsup}
\end{table}%

\begin{table}[t]
    \centering
    \scalebox{0.87}{%
    \tabcolsep=0.1cm
    \begin{tabular}[t]{l ccc ccc ccc}
    \toprule
        & 
        \multicolumn{3}{c}{CIFAR100} &
        \multicolumn{3}{c}{ImageNet100}&
        \multicolumn{3}{c}{CUB} \\
        \cmidrule(lr){2-4} \cmidrule(lr){5-7}  \cmidrule(lr){8-10} 
       setup & All & Old & Novel & All & Old & Novel & All & Old & Novel \\
    \midrule
       Unsup. & 67.4 & 67.0 & 67.6 & 88.1 & 92.2 & 86.8 & 29.5 & 35.2 & 27.6 \\
       GCD~\cite{vaze2022gcd} & 87.9 & 97.8 & 81.2 & 90.2 & 99.2 & 87.2 & 82.6 & 98.5 & 64.0 \\
    \bottomrule
    \end{tabular}%
}
\vspace{-2.5mm}
\caption{
    The $k$NN retrieval performance of ours in Recall@8 on the unsupervised category discovery and the GCD setups
}
\vspace{-3mm}
\label{table:ablation_retrieval}
\end{table}%



\smallbreakparagraph{Comparison with different embedding aggregation.}
We verify the proposed contrastive mean-shift learning by replacing the nearest neighbor retrieval and mean aggregation with random retrieval and learnable attentive aggregation.
For the learnable attentive aggregation, we adopt the cross-attention mechanism~\cite{transformers} without the value projection, using a query embedding as a query and $k$NN embeddings as key and value.
Without value projection, attentive pooling of $k$NN embeddings is analogous to the \textit{attentive} mean shift.
For random aggregation, we randomly select $k$ embeddings for each query image instead of the nearest neighbors, which are then shared among both the original and augmented embeddings for stable learning.

As shown in Table~\ref{table:ablation_attn}, the mean aggregation with $k$NNs is the most effective one.
We observe that the random aggregation significantly deteriorates the performance as training progresses.
On the other hand, the attentive aggregation involves more trainable parameters and exhibits stable learning curves in training but generalizes worse than the mean aggregation method at inference with iterative mean shift, \eg, 2.6\% point drop after iterations on ImageNet100.


\smallbreakparagraph{Unsupervised category discovery with CMS.}
We further analyze our method to interpret its behavior and elucidate why it performs particularly well on GCD by comparing ours on the unsupervised category discovery setup.
As shown in Table~\ref{table:ablation_unsup}, CMS improves clustering effects even in an unsupervised setup but more quickly with additional labels, leveraging the help of higher-quality $k$NNs.
Specifically, when comparing CMS and Vaze~\etal~\citep{vaze2022gcd} without labels, ours better performs by an average of 4.7\%p across three benchmarks.
The gap widens when using additional labels, \ie, on the GCD task, as shown in Table~\ref{table:gcd}.
We then evaluate the quality of the retrieved 8NNs in our method across these setups (Table~\ref{table:ablation_retrieval}), and observe that the model trained on the GCD setup retrieves more accurate embeddings \textit{even for unknown classes} than the one trained on the unsupervised setup.
Since CMS propagates supervision in incorporating the relevant $k$NNs, and the relevant $k$NNs lead robust mean shift as examined in Table~\ref{table:ablation_attn},
it eventually establishes better representations for final clustering.

\section{Conclusion}
We have proposed to revisit the mean-shift algorithm and incorporated it with contrastive representation learning for generalized category discovery.
The training procedure trains an image encoder via contrastive learning of the single-step mean-shifted embeddings.
The evaluation procedure iterates the mean-shift steps, mapping the resultant clustered groups to categories.
While the previous work on GCD often exploits the ground-truth number of classes for clustering,
we avoid using the oracle information and instead estimate the number of clusters using agglomerative clustering. 
In experiments, our method achieves state-of-the-art performance on public GCD benchmarks without bells and whistles.
We believe that the proposed contrastive mean-shift learning will benefit representation learning for other diverse tasks beyond generalized category discovery and image clustering that are addressed in this work.

\smallbreakparagraph{Acknowledgements.}
This work was supported by the NRF grant (NRF-2021R1A2C3012728 ($50\%$)) and the IITP grants (2022-0-00113: Developing a Sustainable Collaborative Multi-modal Lifelong Learning Framework ($45\%$), 2019-0-01906:
AI Graduate School Program at POSTECH ($5\%$)) funded by Ministry of Science and ICT, Korea.
{
    \small
    \bibliographystyle{ieeenat_fullname}
    \bibliography{main}
}

\clearpage
\setcounter{page}{1}
\maketitlesupplementary

\section{Additional details}

\subsection{Uniform and Gaussian kernel on Mean-shift}
We denote a uniform kernel and a Gaussian kernel as $\varphi_{u}$ and $\varphi_{g}$, formulated as follows: 
\begin{align}
    \varphi_{u}(x)&=\begin{cases}
            1 & \text{ if } x \leqslant \delta \\
            0 & \text{ if } x > \delta 
        \end{cases}\label{eq:uniform} \\
    \varphi_{g}(x)&=e^{-\frac{x}{2\sigma^{2}}}\label{eq:gaussian}
\end{align}
where $\delta$ indicates a distance threshold and $\sigma$ indicates a standard deviation for determining the bandwidth of the kernel.
In these way, the conventional mean-shift defines the neighborhood kernels with a fixed radius from the query data point where an arbitrary number of neighborhood data points can be included in $\mathcal{N}(\vv_{i})$.
In contrast, our method sets no limit on the kernel radius yet always exploits a fixed number of neighborhood data points, which is realized with $k$NN search. 
For a fair comparison, we adopt cosine similarity for a distance metric same as our method by reformulating the Eqs.~\ref{eq:uniform} and ~\ref{eq:gaussian} as follows:
\begin{align}
    \vz_{i} &= \frac{\sum_{\vv_j \in \mathcal{N}(\vv_{i})} \varphi(cos(\vv_j,\vv_{i})) \vv_j} {||\sum_{\vv_j \in \mathcal{N}(\vv_{i})} \varphi(cos(\vv_j,\vv_{i})) \vv_j||} \\
    \varphi_{u}(x)&=\begin{cases}
        1 & \text{ if } x \geqslant \delta \\
        0 & \text{ if } x < \delta 
        \end{cases}\label{eq:uniform_cosine} \\
    \varphi_{g}(x)&=e^{-\frac{1-x}{2\sigma^{2}}}\label{eq:gaussian_cosine}
\end{align}
where $cos(\cdot)$ refers to cosine similarity.

\smallbreakparagraph{Implementation details.}
We replace $k$NNs with distance-based NNs with the following implementation details.
We empirically set $\delta=0.9$, $\sigma=0.1$, and limit the maximum number of retrieved embeddings to 1,000 to prevent including too many points within the neighborhood.
During evaluation, we modify the inference process by replacing the fixed $k$NN search with the uniform or Gaussian kernel as well.
For a fair comparison, all the other hyperparamters remain the same as ours.
\label{sec:uniform_gaussian}

\subsection{Details on benchmarks}
Table~\ref{table:split} shows the number of labeled and unlabeled classes of each benchmark.
In Table~\ref{table:ablation_datasize}, we also denote the size of the image collection used for $k$NN retrieval.
\begin{table}
\centering
    \resizebox{0.9\columnwidth}{!}{%
    \begin{tabular}[t]{l | cc}
    \toprule
        & known classes & unknown classes \\
    \midrule
        CIFAR100~\cite{cifar} & 80 & 20 \\
        ImageNet100~\cite{tieredimagenet} & 50 & 50\\
        CUB-200-2011~\cite{cub} & 100 & 100\\
        Stanford-Cars~\cite{krause20133scars} & 98 & 98\\
        FGVC-Aircraft~\cite{maji2013aircraft} & 50 & 50\\
        Herbarium19~\cite{tan2019herbarium} & 341 & 342\\
    \bottomrule
    \end{tabular}%
}
\caption{Number of known and unknown classes.}
\label{table:split}
\end{table}

\begin{table}[t!]
\centering
    \resizebox{\columnwidth}{!}{%
    \begin{tabular}[t]{l c c c c c c}
    \toprule
        & CIFAR100 & ImageNet100 & CUB & Cars & Aircraft & Herbarium \\
    \midrule
        $|\mathcal{D}|$ & 50,000 & 127,115 & 5,994 & 8,144 & 6,667 & 34,225 \\
    \bottomrule
    \end{tabular}%
    }
\caption{
    Size of the $k$NN search space.
}
\label{table:ablation_datasize}
\end{table}%
\label{sec:detail_dataset}

\section{Additional experimental results}

\subsection{Design choices on supervised loss.}
In Table~\ref{table:sup_loss}, we replace the input of $\gL_{\text{SC}}$ with the mean-shifted embedding $z$ as the same as in $\gL_{\text{CMS}}$.
The use of mean-shift in the supervised loss even harms the performance especially on fine-grained benchmarks.
Since CMS learning is designed to incorporate the neighborhood collaboration of the query, 
integrating unlabeled (thus noisy) $k$NNs with the supervised loss turns out to be unreliable.

\begin{table}[t]
    \centering
    \tabcolsep=0.1cm
    \resizebox{\columnwidth}{!}{%
    \begin{tabular}[t]{l ccc ccc ccc ccc}
    \toprule
        \multirow{2.5}{*}{$\gL_{\text{SC}}$} &
        \multicolumn{3}{c}{CIFAR100} &
        \multicolumn{3}{c}{ImageNet100} &
        \multicolumn{3}{c}{CUB} &
        \multicolumn{3}{c}{Stanford Cars} \\
        \cmidrule(lr){2-4} \cmidrule(lr){5-7} \cmidrule(lr){8-10} \cmidrule(lr){11-13}
         & All & Old & Novel & All & Old & Novel & All & Old & Novel & All & Old & Novel\\
    \midrule
        $z$ & 80.1 & \textbf{86.0} & 68.2 & 84.6 & 95.5 & 79.1 & 65.9 & \textbf{82.4} & 62.6 & 40.1 & 62.4 & 29.4 \\
        $v$ & \textbf{82.3} & 85.7 & \textbf{75.5} & \textbf{84.7} & \textbf{95.6} & \textbf{79.2} & \textbf{68.2} & 76.5 & \textbf{64.0} & \textbf{56.9} & \textbf{76.1} & \textbf{47.6} \\
    \bottomrule
    \end{tabular}%
}
\caption{
    Ablation study on different embeddings for $\gL_{\text{SC}}$. The symbol $z$ denotes mean-shifted embedding, and $v$ denotes the initial embedding.
}
\label{table:sup_loss}
\end{table}%

\hspace{2mm}

\begin{table}[t!]
\centering
    \resizebox{0.8\columnwidth}{!}{%
    \begin{tabular}[t]{l ccc ccc}
    \toprule
        \multirow{2.5}{*}{$k$} &
        \multicolumn{3}{c}{CIFAR100} &
        \multicolumn{3}{c}{ImageNet100} \\
        \cmidrule(lr){2-4} \cmidrule(lr){5-7} 
         & All & Old & Novel & All & Old & Novel\\
    \midrule
        4 & 81.7 & 85.4 & 74.3 & 83.2 & 95.5 & 77.0 \\
        8 & \textbf{82.3} & 85.7 & \textbf{75.5} & \textbf{84.7} & 95.6 & \textbf{79.2} \\
        16 & 80.4 & 86.5 & 68.1 & 83.7 & 95.6 & 77.7 \\
        32 & 79.4 & \textbf{86.6} & 65.1 & 83.1 & \textbf{95.7} & 76.7\\
        64 & 77.1 & 84.8 & 61.7 & 83.8 & \textbf{95.7} & 77.8 \\
    \toprule
        \multirow{2.5}{*}{$k$} &
        \multicolumn{3}{c}{Stanford Cars} &
        \multicolumn{3}{c}{CUB}\\
        \cmidrule(lr){2-4} \cmidrule(lr){5-7} 
         & All & Old & Novel & All & Old & Novel\\
    \midrule
        4 & 66.2 & 73.6 & 62.5 & 48.7 & 70.7 & 38.1 \\
        8 & \textbf{68.2} & \textbf{76.5} & \textbf{64.0} & \textbf{56.9} & \textbf{76.1} & \textbf{47.6} \\
        16 & 64.0 & 74.8 & 58.5 & 50.5 & 71.5 & 40.3 \\
        32 & 58.9 & 75.6 & 50.5 & 50.6 & 72.7 & 39.9 \\
        64 & 49.4 & 61.9 & 43.1 & 45.1 & 72.2 & 32.0 \\
    \bottomrule
    \end{tabular}%
    }
\caption{
    Ablation study with varying numbers of nearest neighbors $k$.
}
\label{table:ablation_knn}
\end{table}%
\subsection{Effect of the number of nearest neighbors $k$}
In Table~\ref{table:ablation_knn}, we examine the effect of the number of nearest neighbor $k$ by varying the value with 4, 8, 16, 32, and 64.
The result shows that retrieving 8NNs shows reasonable performance overall.
The higher $k$ value tends to more negatively affect 
on smaller-scale benchmarks such as CUB and Stanford Cars since more unrelated NNs are included due to the smaller size of the search space as outlined in Table~\ref{table:ablation_datasize}.
This also aligns with the experiment on section~\ref{sec:uniform_gaussian}, in the sense that the different number of nearest neighbors can be interpreted as adopting a different bandwidth of a kernel.
In other words, a large number of $k$ leads to a similar outcome of using a larger kernel during the mean-shift clustering, which might over-smooth the embedding space.
This indicates that retrieving an appropriate number of $k$ is essential for stable learning. 

\subsection{Effect of the scaling hyperparameter $\alpha$}
We investigate the effect of the scale parameter $\alpha$ in Table~\ref{table:ablation_alpha}.
As $\alpha$ increases, $k$NN embeddings are assigned higher weights, which can be interpreted as increasing the uniformity of the mean-shift kernel.
Unlike conventional mean-shift algorithms which are often sensitive to kernel parameters, our method demonstrates stable performance across different combinations of queries and $k$NN embeddings.
The method tends to show better performance as $\alpha$ decreases, suggesting that approximating a Gaussian kernel is efficient when well-optimized for the target data distribution.
This optimization is facilitated by the $k$NN-based kernel, which dynamically adjusts the kernel's bandwidth.
Through this, we validate that adopting the mean-shift in a learnable manner ensures consistent shifting, even when a kernel is approximated to a discrete and non-continuous format of the $k$NN retrieval process.

\begin{table}[t]
\centering
    \resizebox{0.8\columnwidth}{!}{%
    \begin{tabular}[t]{l ccc ccc}
    \toprule
        \multirow{2.5}{*}{$\alpha$} &
        \multicolumn{3}{c}{ImageNet100} &
        \multicolumn{3}{c}{CUB} \\
        \cmidrule(lr){2-4} \cmidrule(lr){5-7} 
         & All & Old & Novel & All & Old & Novel\\
    \midrule
        0.5 & \textbf{84.7} & \textbf{95.6} & \textbf{79.2} & \textbf{68.2} & \textbf{76.5} & \textbf{64.0} \\
        0.6 & 83.4 & 95.7 & 77.3 & 63.7 & 74.3 & 58.4 \\
        0.7 & 83.5 & 95.7 & 77.4 & 64.3 & 75.7 & 58.6 \\
        0.8 & 82.4 & 95.7 & 75.6 & 62.7 & 73.8 & 57.2 \\
        1/($k$+1) & 82.1 & 93.9 & 76.2 & 58.7 & 68.4 & 53.9 \\
    \bottomrule
    \end{tabular}%
    }
\caption{
    Ablation study on different choice of scaling hyperparameter $\alpha$ in Eq.~\ref{eq:mean_shifted_embedding_meanvector}
}
\label{table:ablation_alpha}
\end{table}%

\begin{table}[t!]
\centering
    \resizebox{0.8\columnwidth}{!}{%
    \begin{tabular}[t]{l ccc ccc}
    \toprule
        \multirow{2.5}{*}{$\lambda$} &
        \multicolumn{3}{c}{ImageNet100} & 
        \multicolumn{3}{c}{CUB}\\
        \cmidrule(lr){2-4} \cmidrule(lr){5-7}
         & All & Old & Novel & All & Old & Novel \\
    \midrule
        0.25 & 84.7 & 95.5 & 79.3 & 66.8 & 74.9 & 62.8 \\
        0.35 & \textbf{84.7} & \textbf{95.6} & \textbf{79.2} & \textbf{68.2} & \textbf{76.5} & \textbf{64.0}\\
        0.5 & 84.4 & 95.9 & 78.6 & 65.3 & 74.4 & 60.8 \\
    \bottomrule
    \end{tabular}%
    }
\caption{
    Ablation study on the weight of balancing factor $\lambda$ between the losses. 
}
\label{table:ablation_lambda}
\end{table}%

\subsection{Effect of the loss balancing hyperparameter $\lambda$}
In Table~\ref{table:ablation_alpha}, we compare the performance with varying the weights of balancing factor $\lambda$ between the supervised contrastive loss $\gL_{\text{SC}}$ and the unsupervised contrastive mean-shift loss $\gL_{\text{CMS}}$.
The higher the weight, the greater the contribution of labeled images from known classes during training.
Similar to the ablation studies on other hyerparameters, the performance is comparable on ImageNet100, while more sensitive on the fine-grained benchmark, e.g, CUB.
Overall, using larger weight on the supervised loss tends to deteriorates the performance on unknown classes.


\begin{table*}[h!]
    \centering
    \tabcolsep=0.1cm
    \resizebox{2.1\columnwidth}{!}{ 
    \begin{tabular}[b]{l c ccc ccc ccc ccc ccc ccc}
    \toprule
        \multirow{2.5}{*}{Method} &
        \multirow{2.5}{*}{Known $K$} &
        \multicolumn{3}{c}{CIFAR100} &
        \multicolumn{3}{c}{ImageNet100} &
        \multicolumn{3}{c}{CUB} &
        \multicolumn{3}{c}{Stanford Cars} &
        \multicolumn{3}{c}{FGVC Aircraft} &
        \multicolumn{3}{c}{Herbarium 19} \\
        \cmidrule(lr){3-5} \cmidrule(lr){6-8} \cmidrule(lr){9-11}
        \cmidrule(lr){12-14} \cmidrule(lr){15-17} \cmidrule(lr){18-20}
        & & All & Old & Novel & All & Old & Novel & All & Old & Novel
        & All & Old & Novel & All & Old & Novel & All & Old & Novel\\
    \midrule
        GCD~\cite{vaze2022gcd} & \ding{51} & 70.4 & 79.3 & 52.6 & 71.6 & 86.0 & 64.6 & 51.1 & 56.2 & 48.6 
                & 62.5 & 73.9 & 57.0 & 41.2 & 43.0 & 40.2 & \textbf{39.7} & \textbf{58.0} & 29.9 \\
        PromptCAL~\cite{zhang2023promptcal} & \ding{51} & 69.4 & 77.3 & 53.5 & 75.2 & 87.0 & 69.3 & 53.7 & 61.4 & 49.9
                 & 60.1 & 77.9 & 51.5 & 42.2 & 48.4 & 39.0 & 37.4 & 50.6 & \textbf{30.3}\\
        \ccol Ours & \ccol \ding{51} & \ccol \textbf{80.3} & \ccol \textbf{85.2} & \ccol 70.6 & \ccol \textbf{85.9} & \ccol \textbf{93.8} & \ccol \textbf{82.0} & \ccol \textbf{65.8} & \ccol \textbf{75.3} & \ccol 61.1 & \ccol \textbf{77.9} & \ccol \textbf{89.0} & \ccol \textbf{72.6} & \ccol 50.3 & \ccol \textbf{59.1} & \ccol 45.9  & \ccol 36.2 & \ccol 56.5 & \ccol 25.3 \\
        \ccol Ours & \ccol & \ccol 78.0 & \ccol 81.2 & \ccol \textbf{71.5} & \ccol 84.8 & \ccol \textbf{93.8} & \ccol 80.2 & \ccol 65.6 & \ccol 74.0 & \ccol \textbf{61.3} & \ccol 77.2 & \ccol 87.3 & \ccol 72.3 & \ccol \textbf{50.6} & \ccol 52.8 & \ccol \textbf{49.5} & \ccol 38.8 & \ccol 57.7 & \ccol 28.6 \\
    \bottomrule
    \end{tabular}%
    }
    \caption{Comparison of ours and the state of the arts on GCD with CLIP-ViT/B16,
    evaluated \textit{with} or \textit{without} the ground-truth class number $K$ for clustering.
    }
    \label{table:gcd_clip}
\end{table*}%

\begin{table*}[t!]
    \centering
    \tabcolsep=0.13cm
    \resizebox{2.1\columnwidth}{!}{ 
    \begin{tabular}[b]{l c ccc ccc ccc ccc ccc ccc}
    \toprule
        \multirow{2.5}{*}{Method} &
        \multirow{2.5}{*}{Known $K$} &
        \multicolumn{3}{c}{CIFAR100} &
        \multicolumn{3}{c}{ImageNet100} &
        \multicolumn{3}{c}{CUB} &
        \multicolumn{3}{c}{Stanford Cars} &
        \multicolumn{3}{c}{FGVC Aircraft} &
        \multicolumn{3}{c}{Herbarium 19} \\
        \cmidrule(lr){3-5} \cmidrule(lr){6-8} \cmidrule(lr){9-11}
        \cmidrule(lr){12-14} \cmidrule(lr){15-17} \cmidrule(lr){18-20}
        & & All & Old & Novel & All & Old & Novel & All & Old & Novel
        & All & Old & Novel & All & Old & Novel & All & Old & Novel\\
    \midrule
        PromptCAL~\cite{zhang2023promptcal} & \ding{51} & 79.9 & 82.7 & 68.5 & - & - & - & 56.0 & 67.7 & 44.5 
                    & 62.3 & 76.9 & 48.2 & 43.6 & 49.5 & 37.7 & 37.6 & 50.3 & 30.7\\ 
        \ccol Ours & \ccol \ding{51} & \ccol \textbf{80.7} & \ccol 83.9 & \ccol \textbf{68.0} & \ccol \textbf{91.0} & \ccol \textbf{95.9} & \ccol \textbf{86.0} & \ccol \textbf{57.9} & \ccol \textbf{70.2} & \ccol 45.6 & \ccol \textbf{78.2} & \ccol \textbf{88.0} & \ccol \textbf{68.7} & \ccol \textbf{54.1} & \ccol \textbf{59.8} & \ccol \textbf{48.4} & \ccol \textbf{43.0} & \ccol 51.1 & \ccol \textbf{34.5} \\
        \ccol Ours & \ccol & \ccol 80.5 & \ccol \textbf{84.3} & \ccol 65.2 & \ccol 84.7 & \ccol 95.8 & \ccol 73.6 & \ccol 56.1 & \ccol 64.9 & \ccol \textbf{47.4} & \ccol 75.0 & \ccol 87.2 & \ccol 64.2 & \ccol 53.8 & \ccol 59.7 & \ccol 47.9 & \ccol 40.4 & \ccol \textbf{51.3} & \ccol 28.9 \\
    \bottomrule
    \end{tabular}%
    }
    \caption{Comparison of ours and the state of the art on Inductive GCD with CLIP-ViT/B16, 
    evaluated \textit{with} or \textit{without} the ground-truth class number $K$ for clustering. }
    \label{table:inductive_gcd_clip}
\end{table*}%

\begin{table*}[t!]
    \centering
    \centering
    \tabcolsep=0.13cm
    \resizebox{2.1\columnwidth}{!}{ 
    \begin{tabular}[t]{l ccc ccc ccc ccc ccc ccc}
    \toprule
        \multirow{2.5}{*}{Method} &
        \multicolumn{3}{c}{CIFAR100} &
        \multicolumn{3}{c}{ImageNet100} &
        \multicolumn{3}{c}{CUB} &
        \multicolumn{3}{c}{Stanford Cars} &
        \multicolumn{3}{c}{FGVC Aircraft} &
        \multicolumn{3}{c}{Herbarium 19} \\
        \cmidrule(lr){2-4} \cmidrule(lr){5-7} \cmidrule(lr){8-10}
        \cmidrule(lr){11-13} \cmidrule(lr){14-16} \cmidrule(lr){17-19}
         & GT & Pred & Err(\%) & GT & Pred & Err(\%) & GT & Pred & Err(\%) & 
         GT & Pred & Err(\%) & GT & Pred & Err(\%) & GT & Pred & Err(\%) \\
    \midrule
        Ours & 100 & 94 & 6 & 100 & 103 & 3 & 200 & 193 & 3.5 
            & 196 & 189 & 3.6 & 100 & 77 & 23 & 683 & 443 & 35 \\
    \bottomrule
    \end{tabular}%
}
\caption{
    Estimated cluster numbers $K$ and the error rates on the GCD task with CLIP-ViT/B16.
}
\label{table:ablation_clip_kestimation}
\end{table*}%
\subsection{Results with a different image encoder}
We examine the generalizability of our method on different encoders by switching the encoder network from DINO-ViT-B/16 to CLIP-ViT-B/16~\cite{clip}.
For a fair comparison, we reproduce Vaze~\etal~\citep{vaze2022gcd} and PromptCAL~\cite{zhang2023promptcal} using CLIP by fine-tuning the last layer of CLIP and the projection head.
To match the dimension between the encoder and the projection head, 
the input dimension of the projection head is changed from 768 to 512.
The learning rate of PromptCAL~\cite{zhang2023promptcal} is adjusted from 0.1 to 0.01 for improved performance, and all the other hyperparameters remain fixed.

\smallbreakparagraph{Evaluation results on GCD.}
As shown in Table~\ref{table:gcd_clip}, our method on the GCD task outperforms in most cases by a large margin.
Noticeably, the clustering accuracy measured \textit{without} the ground-truth number of $K$ shows comparable performance as well.

\smallbreakparagraph{Evaluation results on inductive GCD.}
We further compare the result on the inductive GCD setup as well.
As shown in Table~\ref{table:inductive_gcd_clip}, our method achieves better accuracy overall, even \textit{without} a given ground-truth number of clusters $K$.

\smallbreakparagraph{Estimated number of clusters.}
In Table~\ref{table:ablation_clip_kestimation}, we report the number of estimated clusters of our method with CLIP encoder.
The reported numbers correspond to the estimated cluster numbers in the experiment shown in Table~\ref{table:gcd_clip}.
This validates the robustness of our method on discovering and estimating clusters with a different image encoder.
\label{sec:clip}


\begin{figure*}[h!]
	\centering
	\small
    \includegraphics[width=1.0\linewidth]{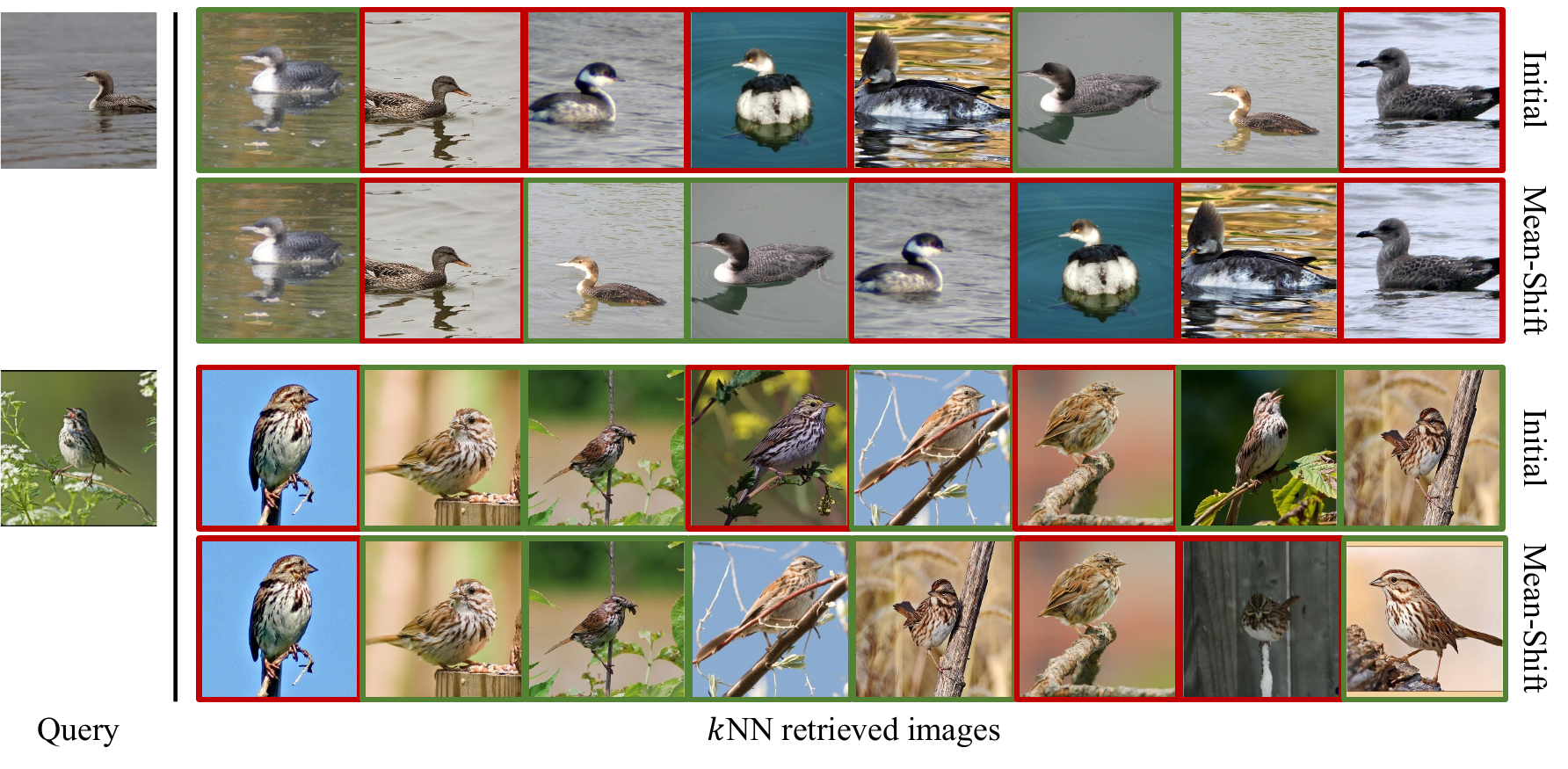}
\vspace{-5mm}
 \caption{$k$NN retrieved images of the initial embedding $\vv$ and mean-shifted embedding $\vz$ on CUB-200-2011. Green denotes the correct class and red an incorrect class.} 
\label{fig:qualitative}
\vspace{-5mm}
\end{figure*}

\vspace{-7mm}
\section{Qualitative results}

\subsection{Qualitative results of the retrieved images.} 
In Figure~\ref{fig:qualitative}, we present $k$NN retrieval results of our model after 1 iteration during inference.
The rows denoted as `Initial' refer to retrieval results using a feature extracted from a trained image encoder as a query.
The rows denoted as `Mean-Shift' indicate using a one-step mean-shifted feature as a query.
The retrieved images are ordered by their similarity scores starting from the top left.
We can observe that applying the mean-shift on learned features enhances the grouping of the instances belonging to the same class, resulting in their retrieval at a higher rank than before.

\begin{figure}[t!]%
    \centering
    \subfloat[DINO~\cite{dino}]{{\includegraphics[width=0.49\columnwidth]{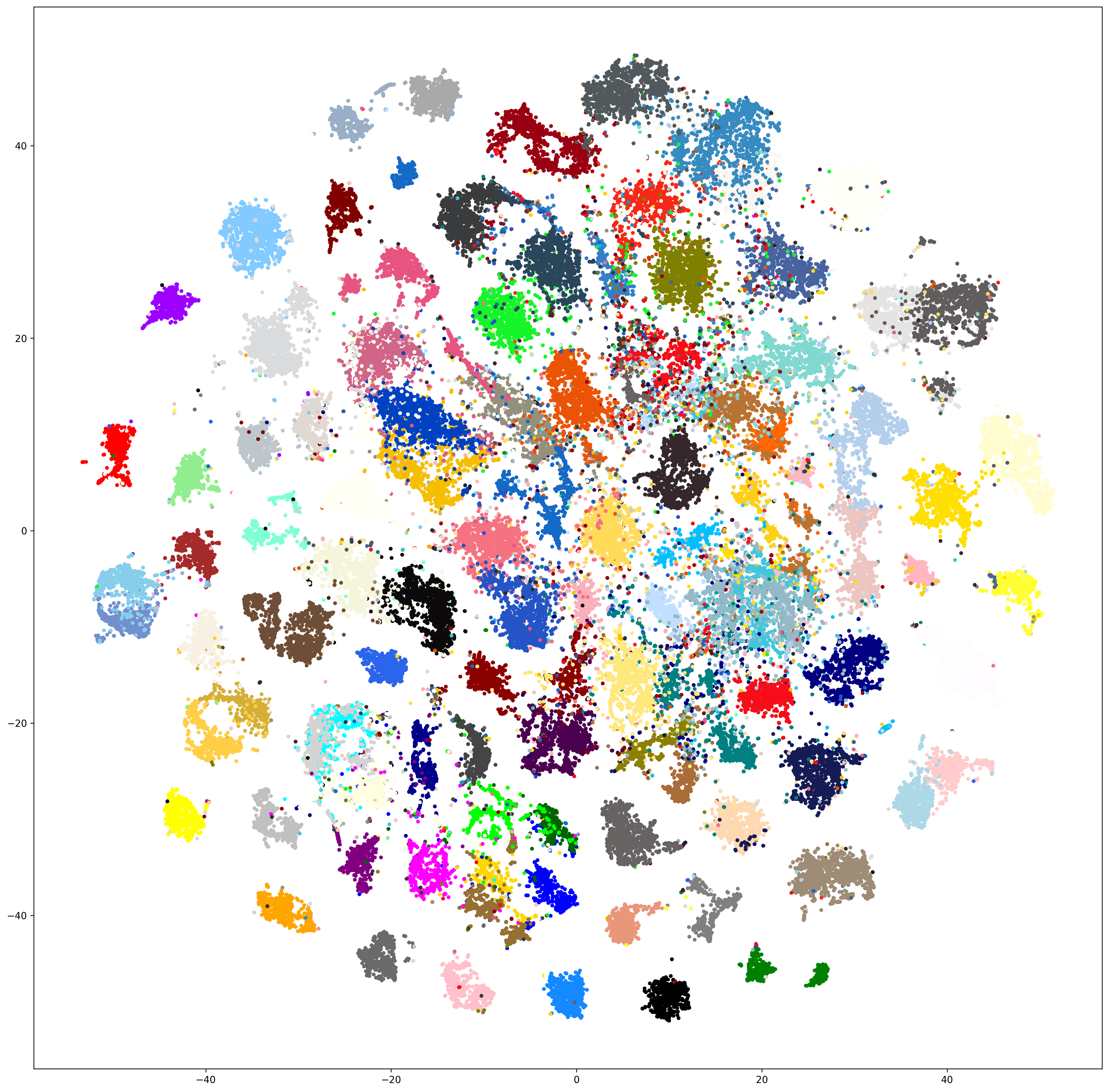}}}%
    \subfloat[Ours]{{\includegraphics[width=0.49\columnwidth]{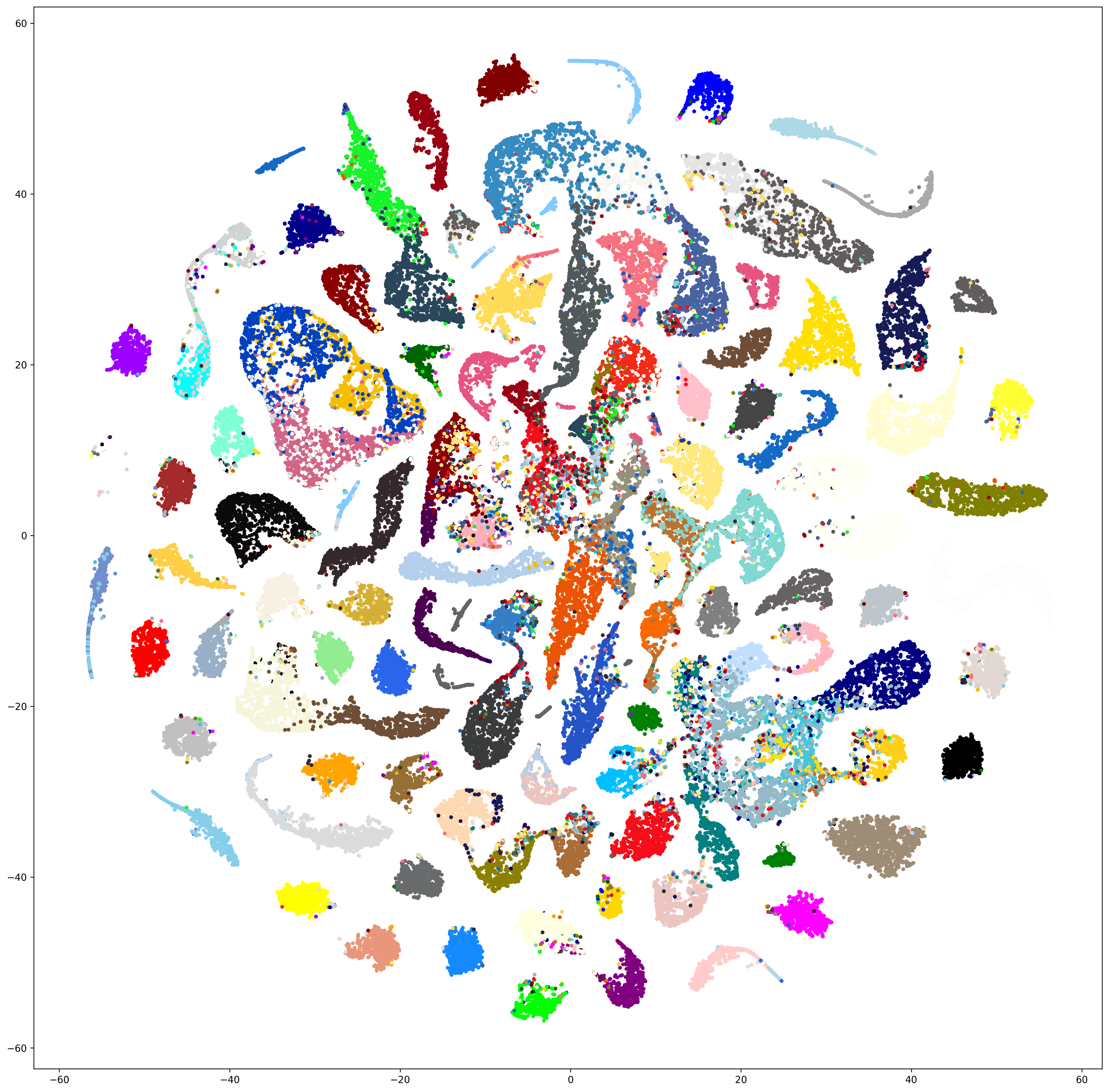}}}%
    \caption{$t$SNE~\cite{tsne} visualization on ImageNet100. Each Color indicates a ground-truth class.}
    \label{fig:tsne}%
\end{figure}

\subsection{tSNE visualization of the embedding space}
In Figure~\ref{fig:tsne}, we visualize the embedding space of ImageNet100 before and after training.
We observe that our method constructs clearer boundaries between clusters.
Notably, the confusing classes are scattered around the center of the plots on the embedding space of DINO, where our method effectively clusters those classes clearer than the DINO baseline once training converged.

\end{document}